\documentclass[11pt]{article}
\usepackage{amsmath}
\usepackage{amssymb}
\usepackage[preprint]{acl}
\usepackage{subcaption}
\usepackage{times}
\usepackage{latexsym}
\usepackage{amsmath}
\usepackage{xcolor}
\definecolor{grassgreen}{HTML}{4E9A06}
\usepackage{booktabs}
\usepackage{arydshln}
\usepackage{multirow}
\usepackage{graphicx}
\usepackage[T1]{fontenc}
\usepackage[utf8]{inputenc}

\usepackage{microtype}

\usepackage{inconsolata}

\usepackage{graphicx}

\title{Revealing the Attention Floating Mechanism in Masked Diffusion Models}
\author{Xin Dai$^{1}$, Pengcheng Huang$^{1}$, Zhenghao Liu$^{1}$\thanks{ \ \ indicates corresponding author.}, Shuo Wang$^{2}$, \\ \textbf{Yukun Yan$^{2}$, Chaojun Xiao$^{2}$, Yu Gu$^{1}$, Ge Yu$^{1}$, Maosong Sun$^{2}$} \\ 
$^1$School of Computer Science and Engineering, Northeastern University, Shenyang, China \\
$^2$Department of Computer Science and Technology, Tsinghua University, Beijing, China\\
}
\begin{document}

\maketitle
\begin{abstract} 
Masked diffusion models (MDMs), which leverage bidirectional attention and a denoising process, are narrowing the performance gap with autoregressive models (ARMs). However, their internal attention mechanisms remain under-explored. This paper investigates the attention behaviors in MDMs, revealing the phenomenon of \textit{Attention Floating}. Unlike ARMs, where attention converges to a fixed sink, MDMs exhibit dynamic, dispersed attention anchors that shift across denoising steps and layers. Further analysis reveals its Shallow Structure-Aware, Deep Content-Focused attention mechanism: shallow layers utilize floating tokens to build a global structural framework, while deeper layers allocate more capability toward capturing semantic content. Empirically, this distinctive attention pattern provides a mechanistic explanation for the strong in-context learning capabilities of MDMs, allowing them to double the performance compared to ARMs in knowledge-intensive tasks. All codes and datasets are available at https://github.com/NEUIR/Attention-Floating.
\end{abstract}

\section{Introduction}

%ARM主流
Large language models (LLMs)~\cite{touvron2023llama,qwen2025qwen25technicalreport} have achieved remarkable success across a wide range of generation and reasoning tasks~\cite{wei2022chain}. The prevailing approach for LLMs has been autoregressive models (ARMs), where a Transformer is trained to predict tokens from left to right. Recent research on ARMs has uncovered a notable \emph{attention sink}~\cite{gu2024attention} phenomenon: a significant portion of the attention mass is systematically absorbed by a few initial tokens at the start of the sequence, which function as prominent static anchors for attention allocation. This rigid attention pattern biases the information flow toward early positions, which can lead to the ``lost-in-the-middle'' issue~\cite{liu2024lost, yao2025spotlight}.

Diffusion language models (DLMs)~\cite{li2022diffusion, liu2023composable, shabalin2025smoothie, shabalin2025tencdm, tae2025tess} have recently emerged as a promising alternative to the autoregressive paradigm, generating text through a multi-step denoising process that relaxes the rigid left-to-right generation constraint in ARMs. As the most prominent instantiation of this paradigm, masked diffusion models (MDMs) begin with a fully masked sequence, then iteratively predict and fill a subset of masked positions across a series of denoising steps. These models employ bidirectional attention, enabling each position to attend to all others. As the sequence evolves from a fully masked state to a fully visible one, its visibility changes across denoising stages.
However, the impact of the gradual denoising process on the attention mechanism remains unclear. Existing work has primarily analyzed attention phenomena in ARMs, such as induction heads~\cite{olsson2022context} and attention sinks~\cite{gu2024attention}, yet the internal workings of MDMs are still largely unexplored.

%贡献总结
% In this paper, we investigate the attention mechanism of MDMs and systematically analyze their attention dynamics. Specifically, we identify the phenomenon of \textit{attention floating} in MDMs, which leads to the dynamic sinking of tokens during attention modeling. Unlike ARMs, where attention rigidly accumulates at a static position (typically the <BOS> token), MDMs exhibit distributed and dynamic structural anchors that shift across denoising steps and layers.
In this paper, we investigate the attention mechanism of MDMs and systematically analyze their attention dynamics. Similar to ARMs, MDMs also exhibit a subset of token positions that receive disproportionately large attention mass. However, unlike ARMs where this behavior often concentrates at a fixed sink at the start of the sequence(typically \texttt{<BOS>}), we identify the phenomenon of \textit{attention floating}: in MDMs, these dominant-attention anchors are dispersed across multiple positions and can shift across denoising steps and layers (Figure~\ref{fig:floating-drift}). We refer to tokens at these positions as \textit{floating tokens}.
Furthermore, we uncover a distinctive mechanism underlying this behavior: \textit{Shallow Structure-Aware, Deep Content-Focused}. Through a geometric decomposition of the pre-softmax attention logits (QK) and an analysis of the specialization of retrieval heads, we demonstrate that the shallow layers rely on structurally floating tokens to form a global framework, while the deeper layers gradually shift attention toward tokens that carry semantic information.

 % Our analysis indicates that the primary reason for this is that the attention floating mechanism uses different tokens to locate more relevant knowledge and is less sensitive to the input positions compared to ARMs~\cite{yao2025spotlight}.

Empirically, we demonstrate that the attention floating mechanism enhances knowledge utilization from in-context inputs. MDMs nearly double the performance gains of ARMs on knowledge-intensive tasks.
To further dissect the underlying drivers of this performance gap, we conduct a series of stress tests to evaluate the models' resilience against contextual noise, positional biases, and complex evidence layouts. These tests reveal that MDMs maintain greater stability in diverse configurations than ARMs~\cite{yao2025spotlight}.
Finally, we employ region-level attention flow analysis, a coarse-grained visualization that tracks how attention mass moves between regions of the input (e.g., \texttt{<BOS>}, query, and evidence), to uncover the underlying mechanism: unlike ARMs that remain rigidly anchored at the sequence start, MDMs dynamically reorganize their internal information pathways to track relevant context actively. These findings provide a mechanistic explanation for the superior robustness of MDMs under in-context learning.

\begin{figure}[t]
  \includegraphics[width=\columnwidth]{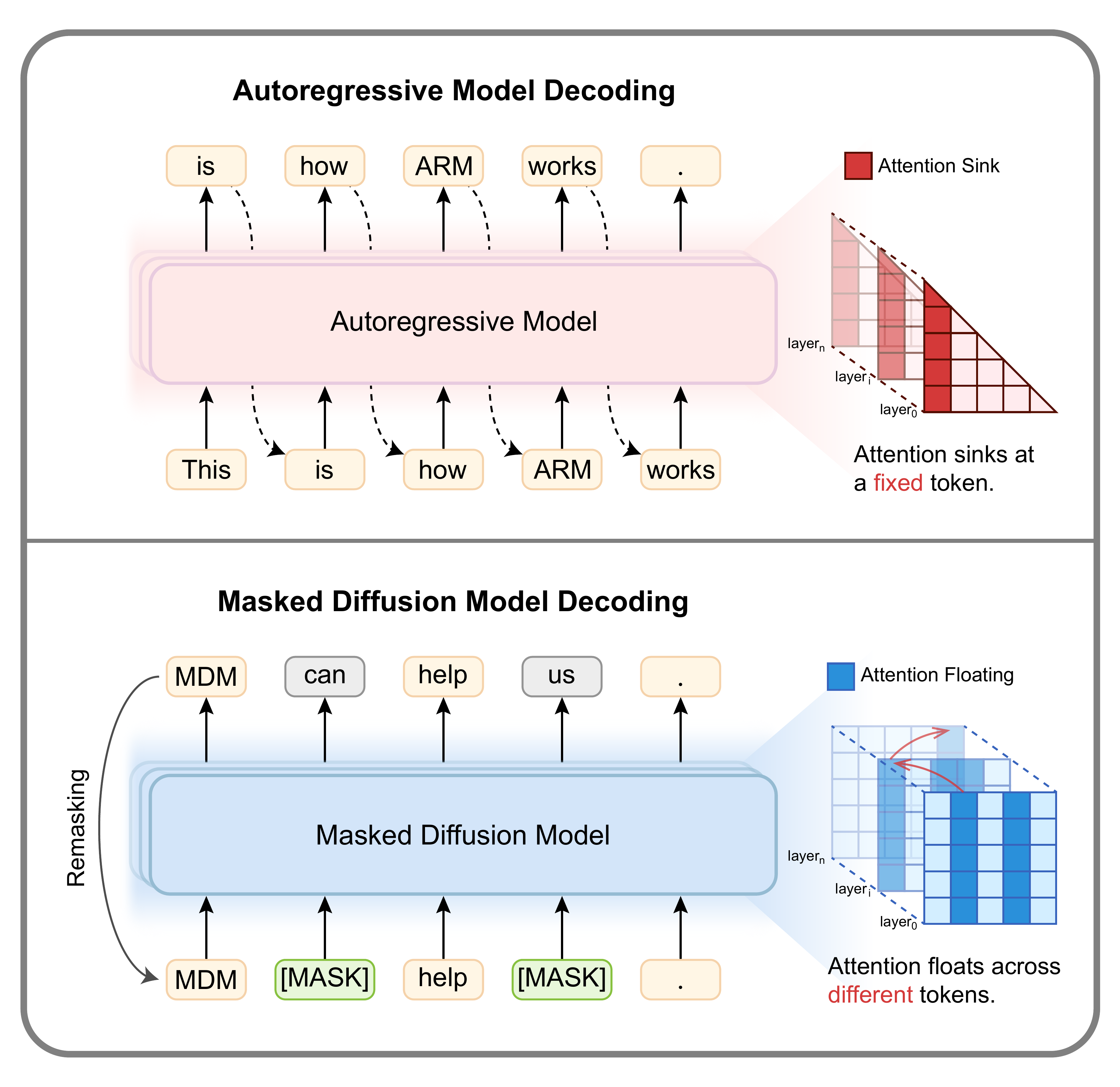}
  \caption{Comparison of ARMs and MDMs.}
  \label{fig:architecture}
\end{figure}

\section{Background: Generative Paradigms and Attention Mechanisms}
In this section, we provide the background for our analysis. We first clarify the architectural differences between autoregressive models (ARMs) and masked diffusion models (MDMs), which motivate our investigation into their attention mechanisms. Subsequently, we introduce the attention sink definition and phenomenon in ARMs, which serves as a comparative baseline for characterizing the ``attention floating'' phenomenon in MDMs.

\subsection{Generative Paradigms: From Autoregressive Decoding to MDMs}
To understand the attention mechanisms in LLMs, we begin by contrasting the architectural differences between autoregressive models (ARMs) and masked diffusion models (MDMs) in their decoding paradigms.

As shown in Figure~\ref{fig:architecture}, the dominant paradigm, ARMs, generate text through a strict left-to-right sequential prediction process. This relies on causal self-attention, where each token is restricted to attending only to its predecessors. In contrast, MDMs employ bidirectional attention, enabling each token to attend to all other positions in the sequence, regardless of their order, at every denoising step. Starting from a fully masked sequence, MDMs conduct denoising steps to recover the target response during inference. Specifically, they progressively reconstruct the text over diffusion time steps, updating multiple tokens in a single step, which facilitates parallel generation.
The transition from causal to bidirectional attention fundamentally alters the flow of information within LLMs. Unlike ARMs, which face limitations such as restricted receptive fields, MDMs provide a global context view. However, it remains unclear how this bidirectional visibility, coupled with time-dependent denoising, impacts the model's attention patterns. To explore this, we first provide background on existing analyses of attention mechanisms in Section~\ref{3.2}.

\subsection{Attention Analysis in Language Models}
\label{3.2}
A large body of prior work has investigated how to analyze attention mechanisms in language models in order to identify the most influential parts of the input~\cite{clark2019does}. With the emergence of LLMs, research has predominantly focused on ARMs. Recent studies show that ARMs exhibit a distinctive \emph{attention sink} phenomenon~\cite{gu2024attention}, which has been widely analysed and exploited in practical settings such as streaming inference and long-context generation.

Specifically, in Transformer-based ARMs, left-to-right dependencies are enforced via self-attention with a causal mask. When a token sequence $X=\{x_1, \dots, x_n\}$ is fed into the model, ARMs encode it using the self-attention mechanism. At the $l$-th layer, the attention weight from the $i$-th token to the $j$-th token is obtained by averaging the attention scores over the $m$ attention heads of ARMs:
\begin{equation}
\small
    A^{\ell}_{i \rightarrow j} = \frac{1}{m} \sum_{h=1}^m \mathrm{Softmax}\!\left(
        \frac{Q^{(\ell,h)}_i K_j^{(\ell,h)\top}}{\sqrt{d_h}}
    \right),
\end{equation}
where $Q^{(\ell,h)}$ and $K^{(\ell,h)}$ denote the query and key matrices of the $h$-th attention head at layer $\ell$, and $d_h$ is the head dimension. 
Then, the average attention received by position $j$ at layer $\ell$ is computed by taking the mean of the attention weights over all $n$ input tokens $X$:
\begin{equation}\label{eq:attn-concentration}
\small
    A^{\ell}_j = \frac{1}{n} \sum_{i=1}^n  A^{\ell}_{i \rightarrow j},
\end{equation}
Finally, the $j$-th token is regarded as the sink token if it satisfies the following criteria:
\begin{equation}\label{eq:attn-sink}
\small
A^{\ell}_j >
\frac{1}{n-1}
\sum_{\substack{k=1 \\ k \neq j}}^{n} A^{\ell}_k
+ \epsilon,
\end{equation}
where $k$ ranges over all positions in the input sequence $X$ and $\epsilon$ is a predefined threshold. 
In ARMs, early tokens are visible to all subsequent positions under the causal mask, making them natural global sinks, so the \texttt{<BOS>} token frequently dominates in practice. 
In contrast, MDMs adopt bidirectional attention mechanisms. It remains unclear whether the attention sink phenomenon persists under such attention patterns, which motivates our further investigation.

\begin{figure}[t]
    \centering

    % ================== Row 1: Case 1 (long) ==================
    \begin{subfigure}[t]{0.49\linewidth}
        \centering
        \includegraphics[width=\linewidth]{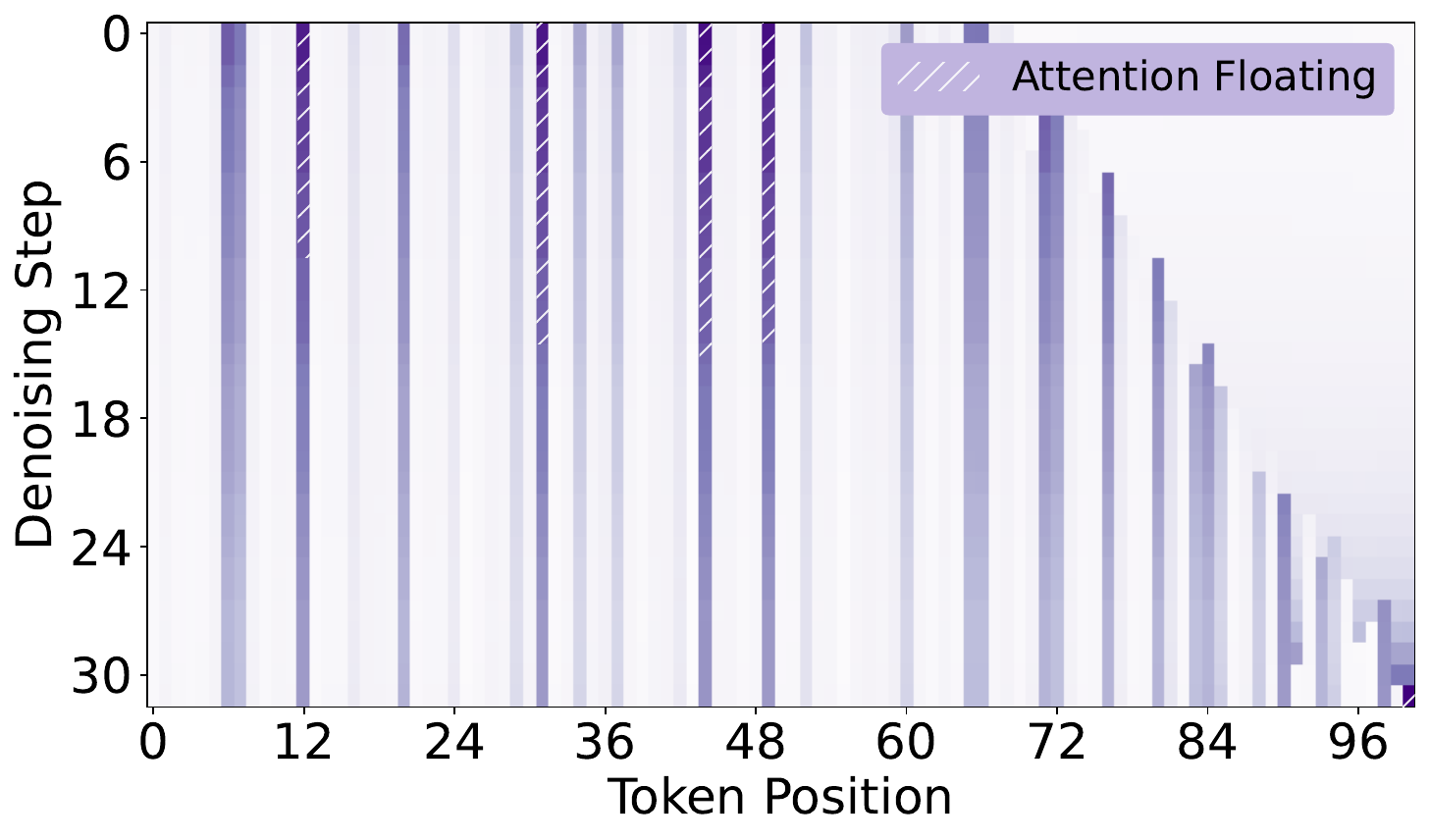}
        \caption{Shallow Layer (Layer 0).}
        \label{fig:floating-case1-layer0}
    \end{subfigure}\hfill
    \begin{subfigure}[t]{0.49\linewidth}
        \centering
        \includegraphics[width=\linewidth]{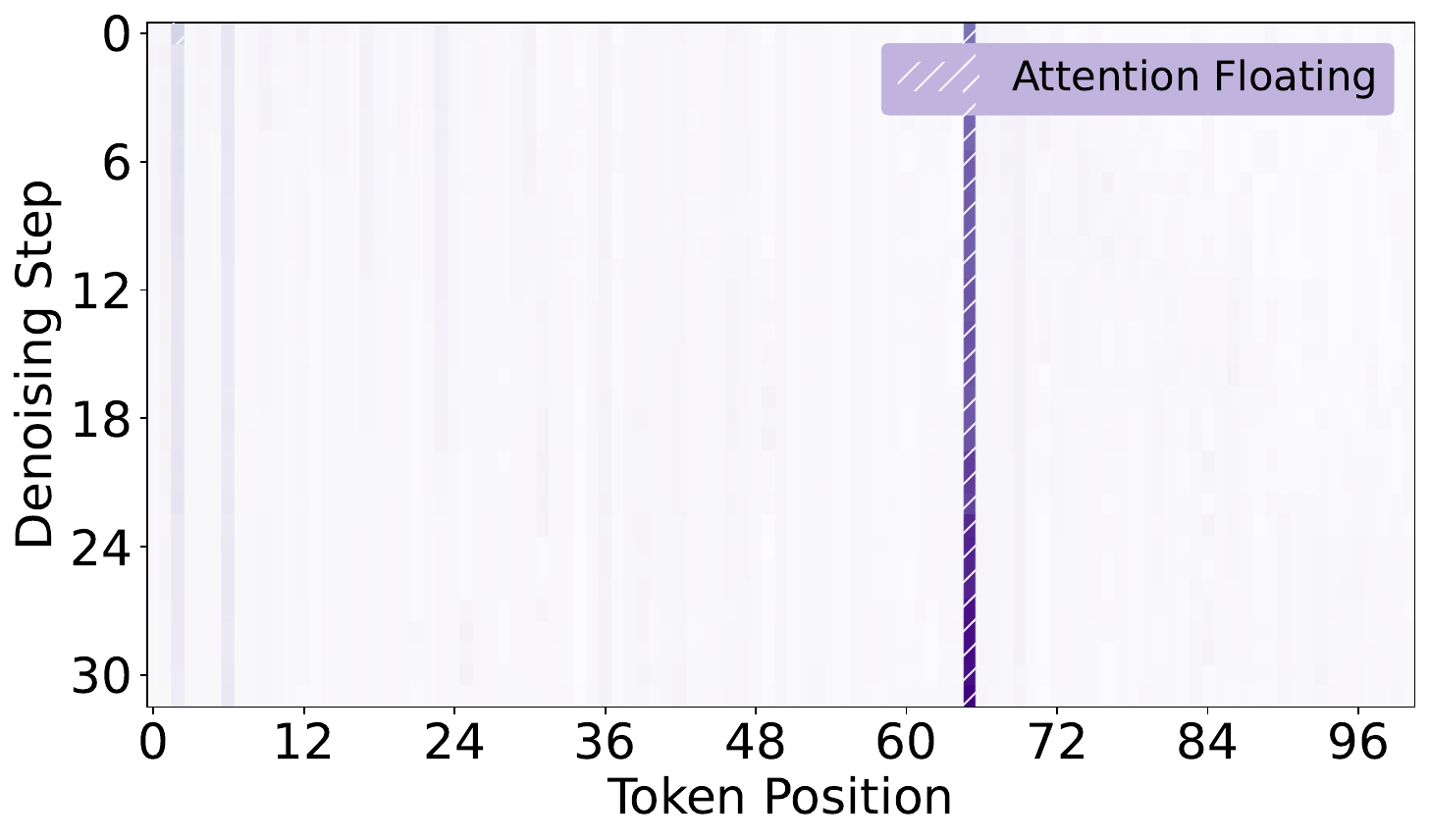}
        \caption{Deep Layer (Layer 31).}
        \label{fig:floating-case1-layer31}
    \end{subfigure}

    \caption*{Case~1 on GSM8K Dataset.}

    \vspace{0.7em}

    % ================== Row 2: Case 2 (standard) ==================
    \begin{subfigure}[t]{0.49\linewidth}
        \centering
        \includegraphics[width=\linewidth]{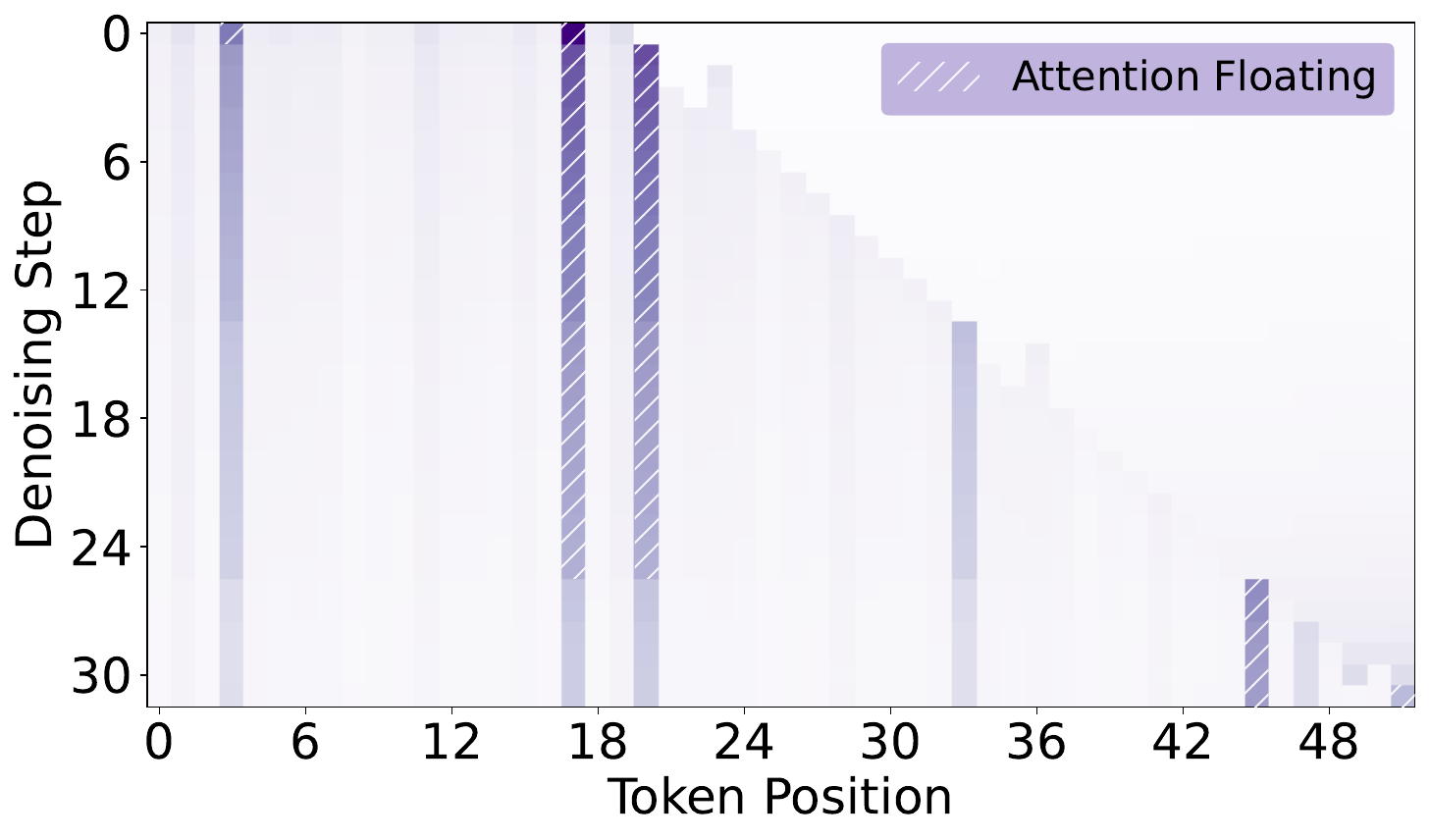}
        \caption{Shallow Layer (Layer 0).}
        \label{fig:floating-case2-layer0}
    \end{subfigure}\hfill
    \begin{subfigure}[t]{0.49\linewidth}
        \centering
        \includegraphics[width=\linewidth]{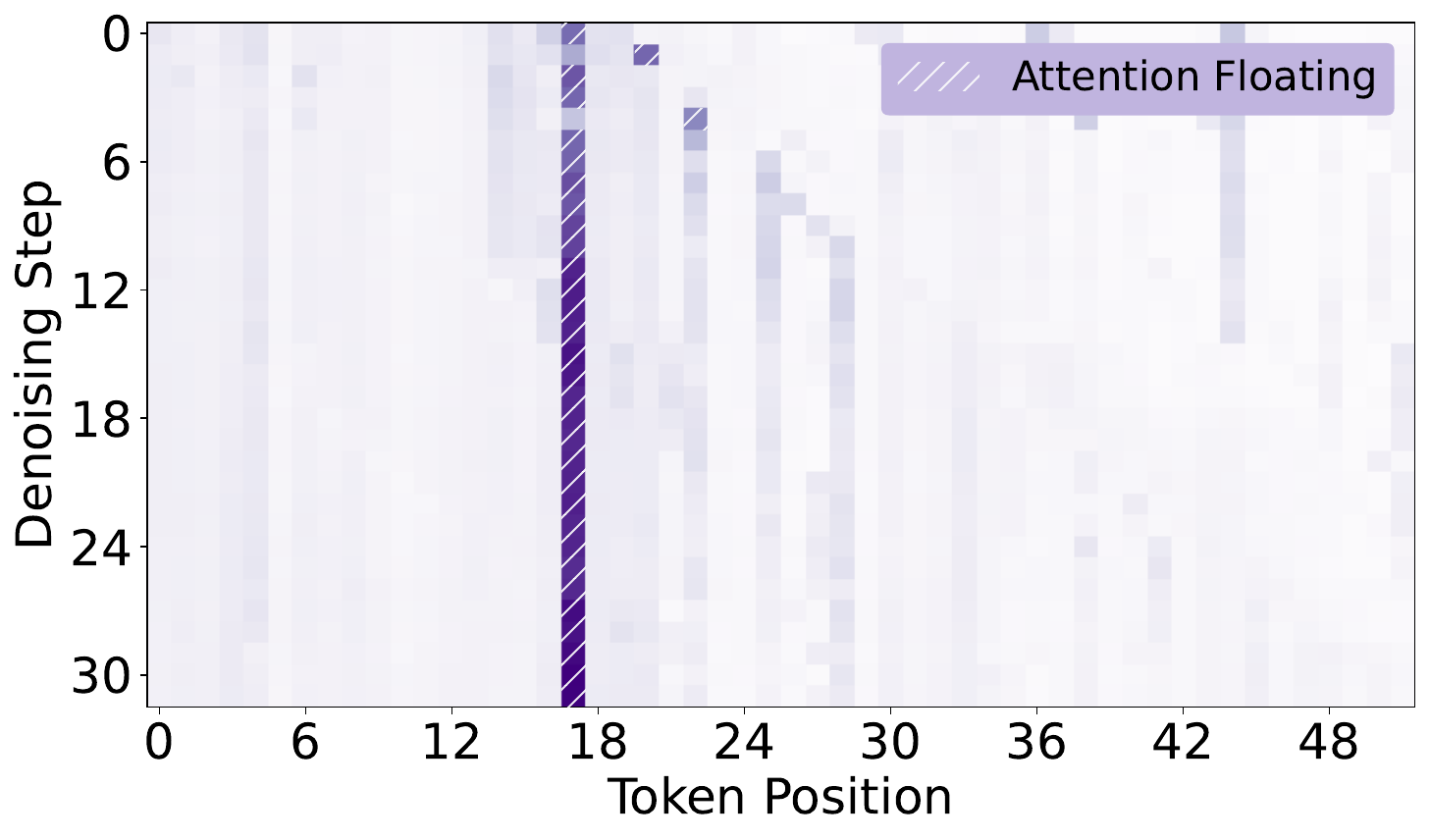}
        \caption{Deep Layer (Layer 31).}
        \label{fig:floating-case2-layer31}
    \end{subfigure}

    \caption*{Case~2 on 2wiki Dataset.}

    \caption{
    Positional Drift of Attention Floating across Different Layers and Denoising Steps in MDM.
    }
    \label{fig:floating-drift}
\end{figure}

\section{Attention: Floating Rather than Sinking in MDMs}
\label{subsec:attn-floating-properties}
Following the attention sink phenomenon observed in ARMs, this section further investigates the attention mechanism of MDMs and reveals the presence of \emph{attention floating}. We then analyse the category of tokens that receive a larger attention weight (floating tokens) in MDMs.

\subsection{Attention Floating Phenomenon in MDMs}
To characterize how attention evolves across different layer depths and denoising steps, we analyze the attention floating mechanism of MDMs through attention visualization.

\textbf{Attention Floating Visualization.}
To illustrate the attention behavior in MDMs, we visualize the per-token attention weights over the input sequence $X$ from the shallow layer ($A^{\ell=0}_j$) to the deep layer ($A^{\ell=31}_j$), computed using Eq.~\ref{eq:attn-concentration}.

In MDMs, we observe an \emph{attention floating} phenomenon, where positions receiving large attention mass shift across layers and further drift as denoising progresses. Moreover, this pattern exhibits task-dependent variations.
Figure~\ref{fig:floating-drift} compares two representative cases on GSM8K and 2WikiMQA and shows that the \emph{attention floating} phenomenon is consistently present across both layer and task.
Across layers, the floating tokens concentrate at clearly different token positions: in the shallow layer (Figure~\ref{fig:floating-case1-layer0} and Figure~\ref{fig:floating-case2-layer0}), high-attention columns spread across multiple positions and gradually shift toward later token positions as the denoising step increases (from around position~$17$ before step~$14$ to position~$34$ after step~$14$, and reaching position~$46$ after step~$26$ in Figure~\ref{fig:floating-case2-layer0}), whereas in the deep layer (Figure~\ref{fig:floating-case1-layer31} and Figure~\ref{fig:floating-case2-layer31}) they become much sparser and concentrate on different positions.
% This layer-wise shift indicates that floating is not a fixed sink at the sequence start, but a set of salient positions whose locations vary with depth.
Moreover, in the same layer, the floating positions differ across tasks: Case~1 and Case~2 exhibit high-attention columns at different token positions, suggesting that floating adapts to task-specific input structure rather than remaining fixed at a single position.
We provide additional visualizations across all layers for MDMs in Figure~\ref{fig:temporal_all} and Figure~\ref{fig:temporal_dream} (Appendix~\ref{drift}).

These observations motivate two complementary analyses along layers and task settings.
Section~\ref{sec4} takes a layer-wise view to decompose attention behavior in MDMs.
Section~\ref{sec5} takes a task-wise view to further examine the link between attention floating and robustness in learning from context.
% In lower layers (Figure~\ref{fig:floating-layer0}), we observe a pronounced positional drift phenomenon.
% Specifically, multiple vertical bands exhibit significantly higher incoming attention, and the identified
% floating tokens form bundles of vertical trajectories whose centers gradually shift toward later
% token positions as the denoising step increases, with attention mass shared across tokens instead of concentrating on a single position.
% In contrast, in deeper layers (Figure~\ref{fig:floating-layer31}), high-attention floating tokens become much sparser, and their positions remain largely stable across denoising steps.
% This suggests that attention floating exhibits stronger positional drift in shallow
% layers and progressively stabilizes in deeper layers. We provide additional visualizations of positional drift across all layers for MDMs in Figure~\ref{fig:temporal_all} and Figure~\ref{fig:temporal_dream} (Appendix~\ref{drift}).

\textbf{Attention Absorption Rate.} We then identify the positions $\mathcal{S}$ of the tokens receiving dominant attention weights using Eq.~\ref{eq:attn-sink}, and use them to quantify how sink tokens in ARMs or floating tokens in MDMs collectively absorb attention.

\begin{figure}[t]
  \includegraphics[width=\columnwidth]{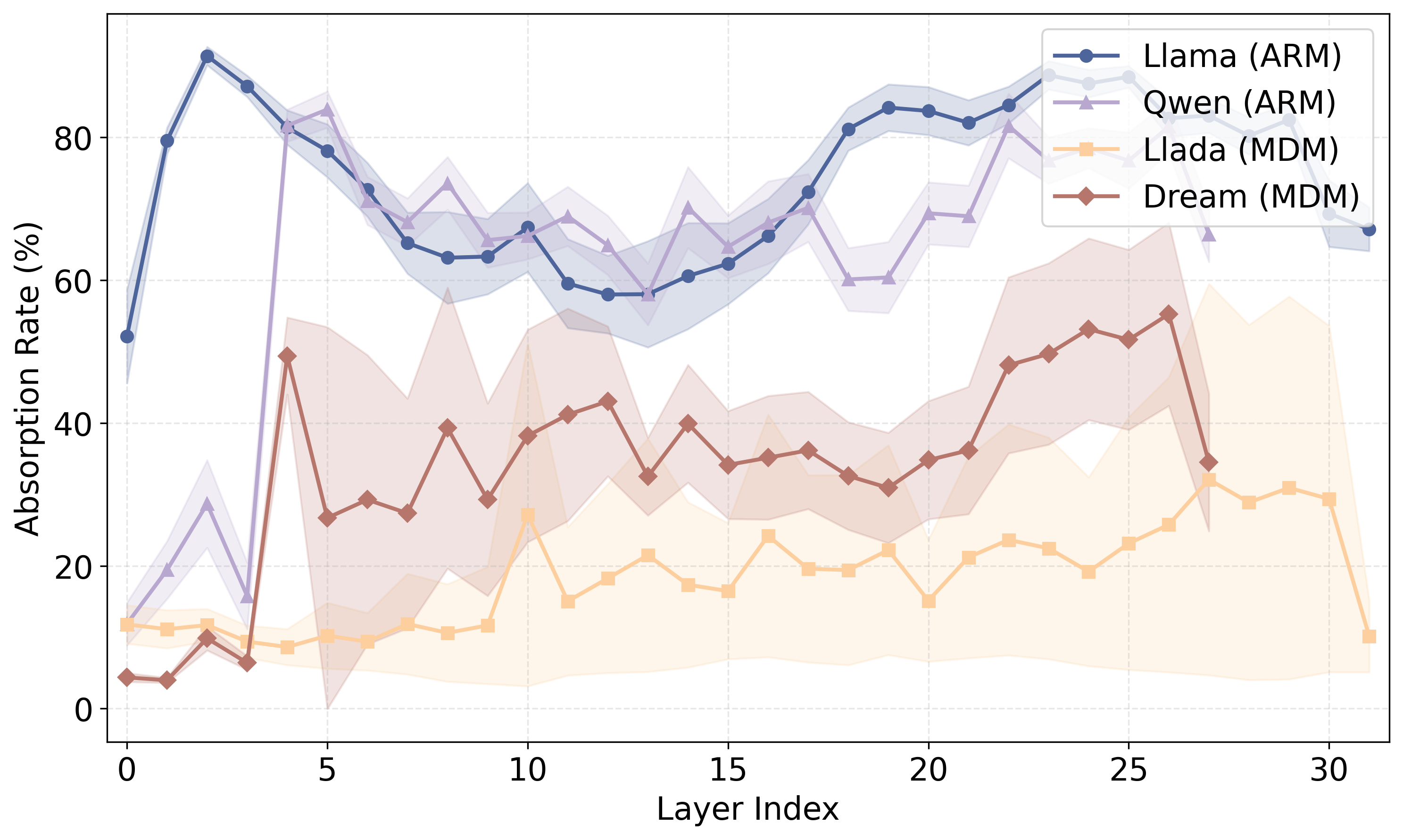}
  \caption{Layer-Wise Attention Absorption Rate in ARMs and MDMs.}
  \label{fig:absorb}
\end{figure}

Specifically, we define the attention absorption rate of the $\ell$-th layer as:
\begin{equation}
\small
\mathrm{Absorb}(\mathcal{S}, \ell) = \sum_{j \in \mathcal{S}} A_j^\ell \times 100\%
\end{equation}
where $A_j^\ell$ denotes the head-averaged attention received by position $j$ at layer $\ell$, and $\mathcal{S}$ represents the position set of sink or floating tokens.
Figure~\ref{fig:absorb} presents the absorption rates of typical ARMs (Llama and Qwen) and MDMs (Llada and Dream) across different layers. Both ARMs use a single \texttt{<BOS>} token as the sink token and exhibit an extremely strong attention sink phenomenon. The sink token absorbs a disproportionately large fraction of the total attention mass across layers. In contrast, MDMs yield lower absorption values. This clear discrepancy indicates that ARMs induce a rigid concentration of attention around the sink token, whereas MDMs display a weaker and more distributed absorption pattern, which aligns precisely with our notion of \emph{attention floating}.

\subsection{Which Tokens Become Floating Tokens}
\label{structure}
Beyond positional drift, we further investigate which types of tokens are prone to becoming floating tokens in MDMs.

For each identified floating token, we classify its underlying vocabulary item into two categories:
(i) structural tokens, including special control tokens (e.g., \texttt{<BOS>}, \texttt{<|mdm\_mask|>}) as well as conventional formatting tokens such as punctuation and other layout symbols; and (ii) lexical tokens, covering content words and subwords. Unlike ARMs, where the attention sink phenomenon is largely dominated by a single special token \texttt{<BOS>}, as shown in Table~\ref{tab:floating_tokens_freq}, floating tokens in MDMs are predominantly composed of high-frequency structural tokens. Within this structural category, approximately 2\% of all detected floating tokens correspond to the model-specific denoising mask token \texttt{<|mdm\_mask|>}. These tokens function less as carriers of semantic content and more as ``structural controllers'' that signal text boundaries and layout organization, thereby anchoring local context and stabilizing the overall sequence structure. The frequency breakdown of floating tokens is summarized in Appendix~\ref{High-Frequency}.

\section{Shallow Structure–Aware, Deep Content–Focused Attention in MDMs}
\label{sec4}
In this section, we take a layer-wise, model-level view of attention in MDMs. We analyze how geometric factors of attention vary across layers, leading to the hypothesis that attention floating shifts from structural bias in shallow layers to semantic content bias in deeper layers, and we then verify this hypothesis by locating retrieval-specialized heads across depth and measuring their contribution to context-following behavior.

\textbf{Self-Attention Decomposition.}
To systematically analyze attention floating in MDMs, we start from the QK scoring mechanism analysis. 

Existing work~\cite{gu2024attention} on ARMs has revealed that the salience of sink positions is primarily manifested as a systematic advantage in the directional (angular) term, while column-wise differences in the scale (norm product) term are relatively weak, and this phenomenon can be summarized as a form of \emph{key bias}.
To disentangle the effect of vector magnitude from directional alignment in the representation space of MDMs, we explicitly decompose the QK score as follows:
\begin{equation}
\small
\mathbf{Q}\mathbf{K}^\top \;=\; \|\mathbf{Q}\|\,\|\mathbf{K}\|\cos\theta,
\end{equation}
where \(\mathbf{Q}\) and \(\mathbf{K}\) denote the query and key vectors and \(\theta\) is the angle between them. This decomposition separates the contribution of vector norms \(\|\mathbf{Q}\|\,\|\mathbf{K}\|\) from that of angular alignment \(\cos\theta\).
%新实验

\begin{figure}[t]
    \centering

    % ---------- Row 1: shallow (Layer 3) ----------
    \begin{subfigure}[t]{0.32\linewidth}
        \centering
        \includegraphics[width=\linewidth]{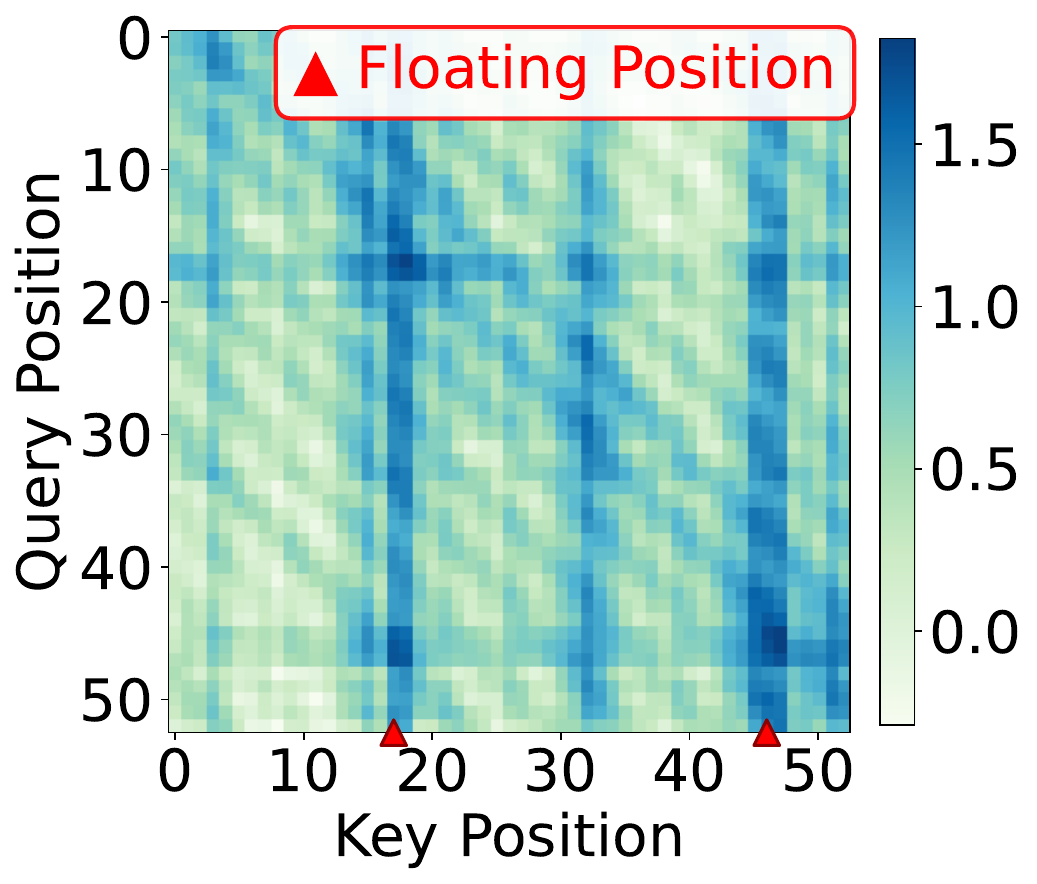}
        \caption{QK Score.}
        \label{fig:qk_decomp_l3_qk}
    \end{subfigure}\hfill
    \begin{subfigure}[t]{0.32\linewidth}
        \centering
        \includegraphics[width=\linewidth]{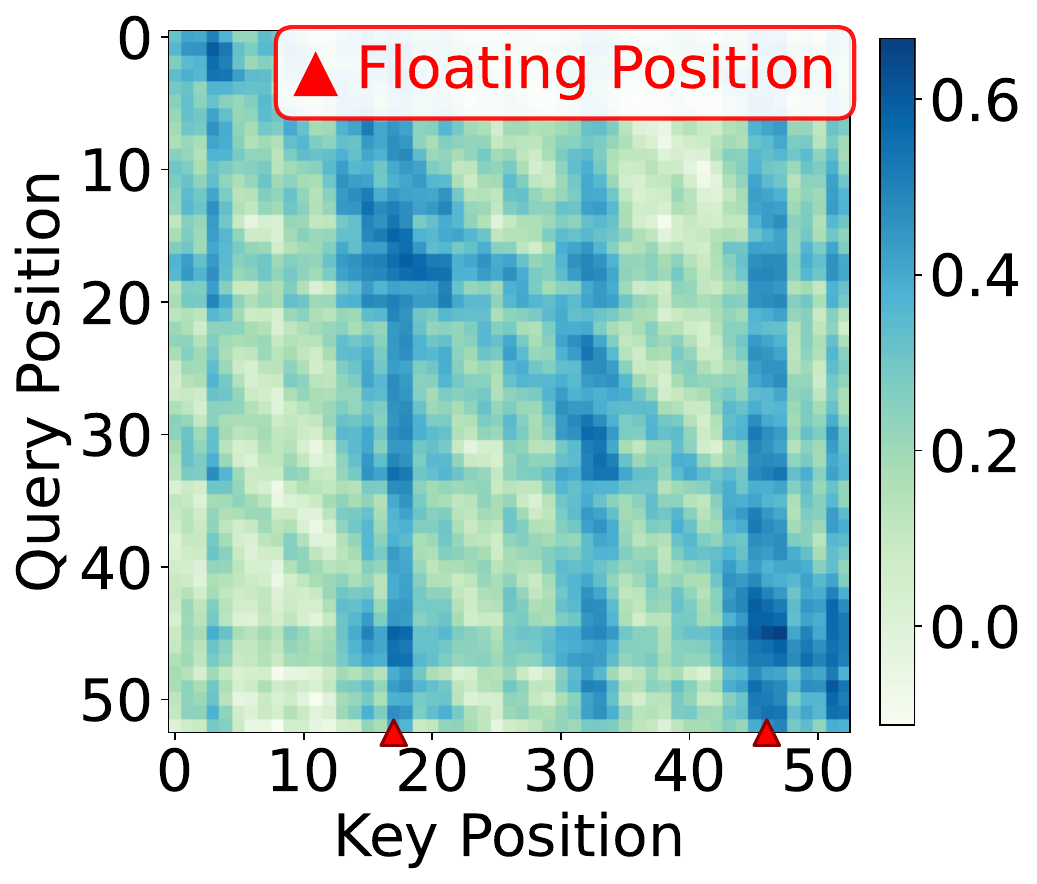}
        \caption{Angular.}
        \label{fig:qk_decomp_l3_cos}
    \end{subfigure}\hfill
    \begin{subfigure}[t]{0.32\linewidth}
        \centering
        \includegraphics[width=\linewidth]{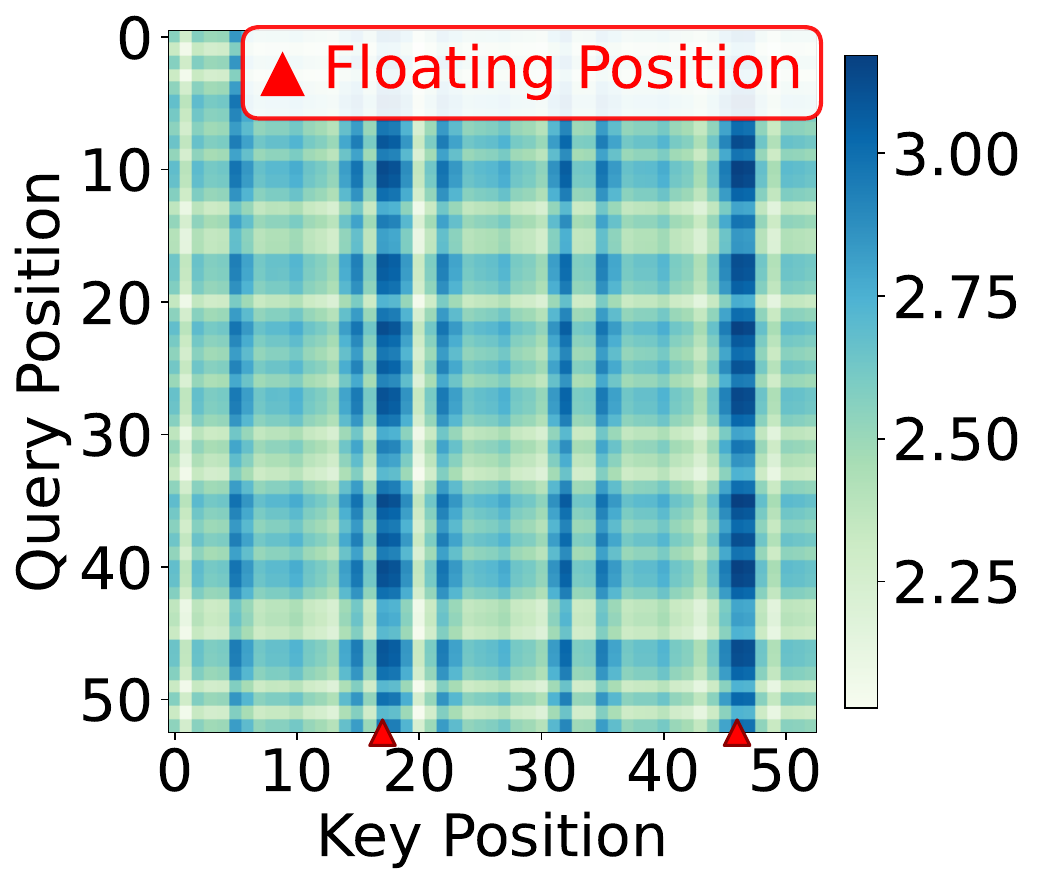}
        \caption{Norm Product.}
        \label{fig:qk_decomp_l3_norm}
    \end{subfigure}

    \vspace{2pt}
    {\small\textit{Layer 1.}}\par
    \vspace{6pt}

    % ---------- Row 2: middle (Layer 11) ----------
    \begin{subfigure}[t]{0.32\linewidth}
        \centering
        \includegraphics[width=\linewidth]{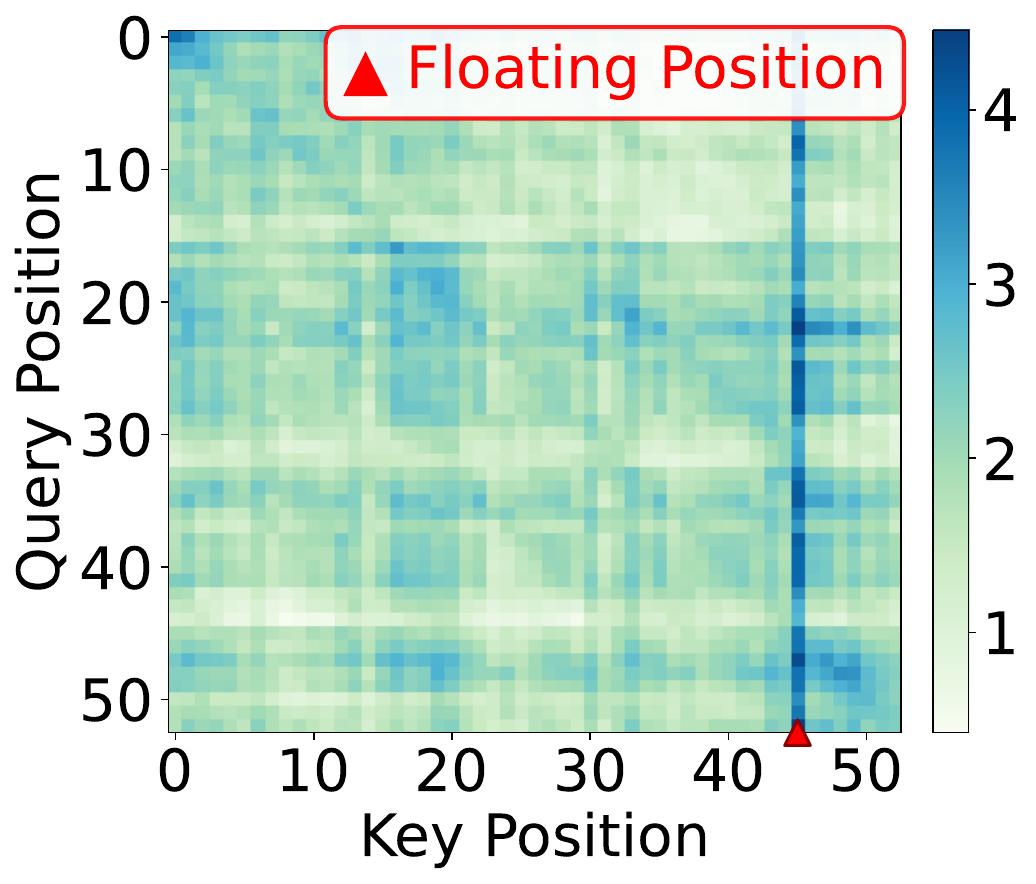}
        \caption{QK Score.}
        \label{fig:qk_decomp_l11_qk}
    \end{subfigure}\hfill
    \begin{subfigure}[t]{0.32\linewidth}
        \centering
        \includegraphics[width=\linewidth]{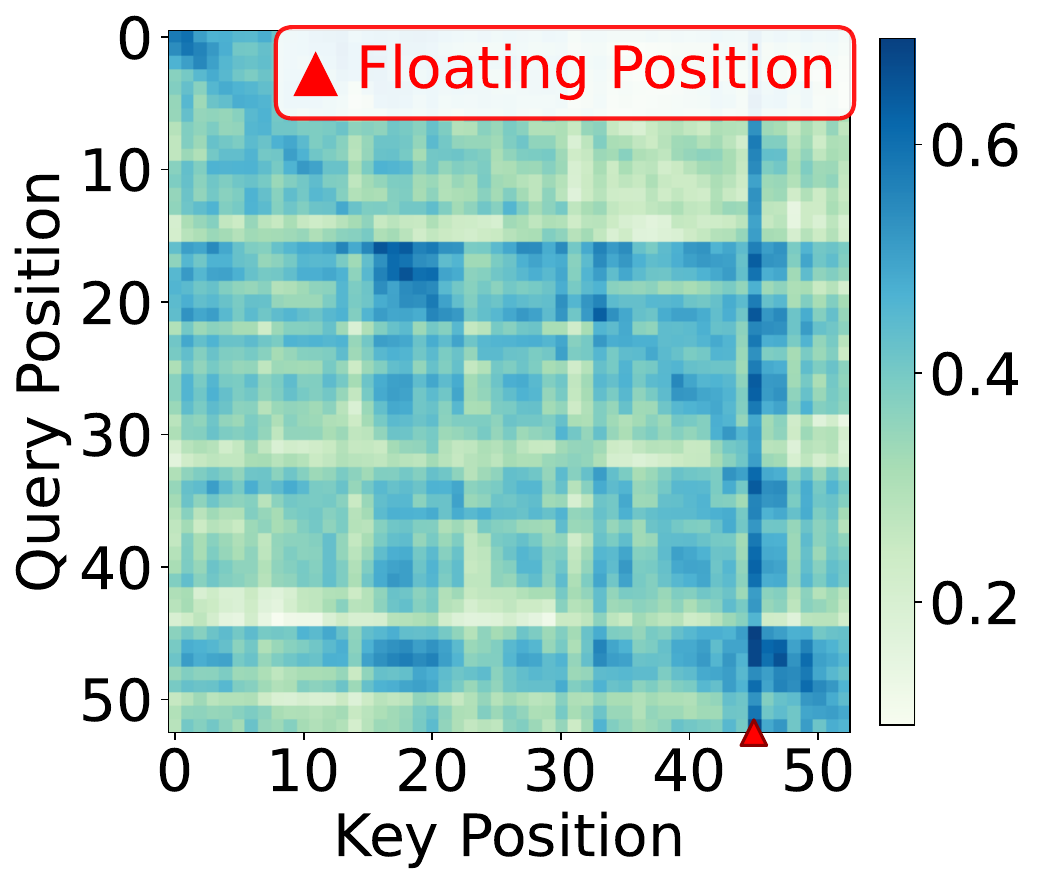}
        \caption{Angular.}
        \label{fig:qk_decomp_l11_cos}
    \end{subfigure}\hfill
    \begin{subfigure}[t]{0.32\linewidth}
        \centering
        \includegraphics[width=\linewidth]{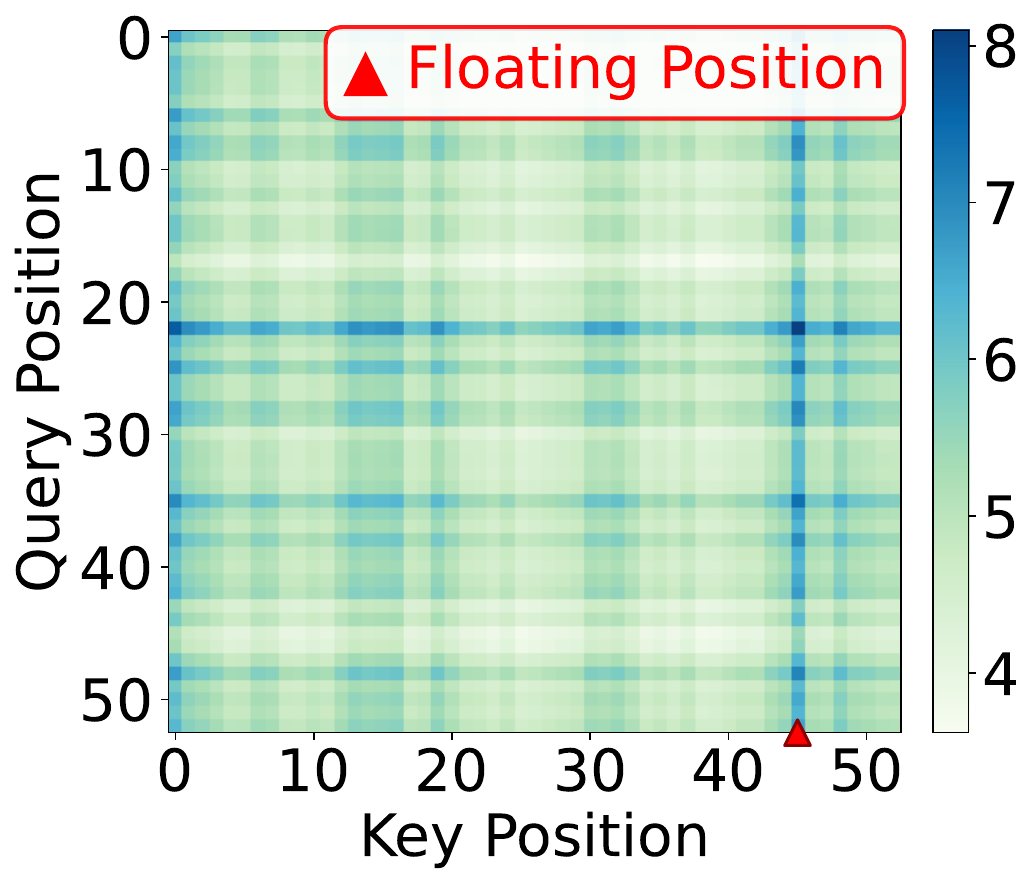}
        \caption{Norm Product.}
        \label{fig:qk_decomp_l11_norm}
    \end{subfigure}

    \vspace{2pt}
    {\small\textit{Layer 11.}}\par
    \vspace{6pt}

    % ---------- Row 3: deep (Layer 28) ----------
    \begin{subfigure}[t]{0.32\linewidth}
        \centering
        \includegraphics[width=\linewidth]{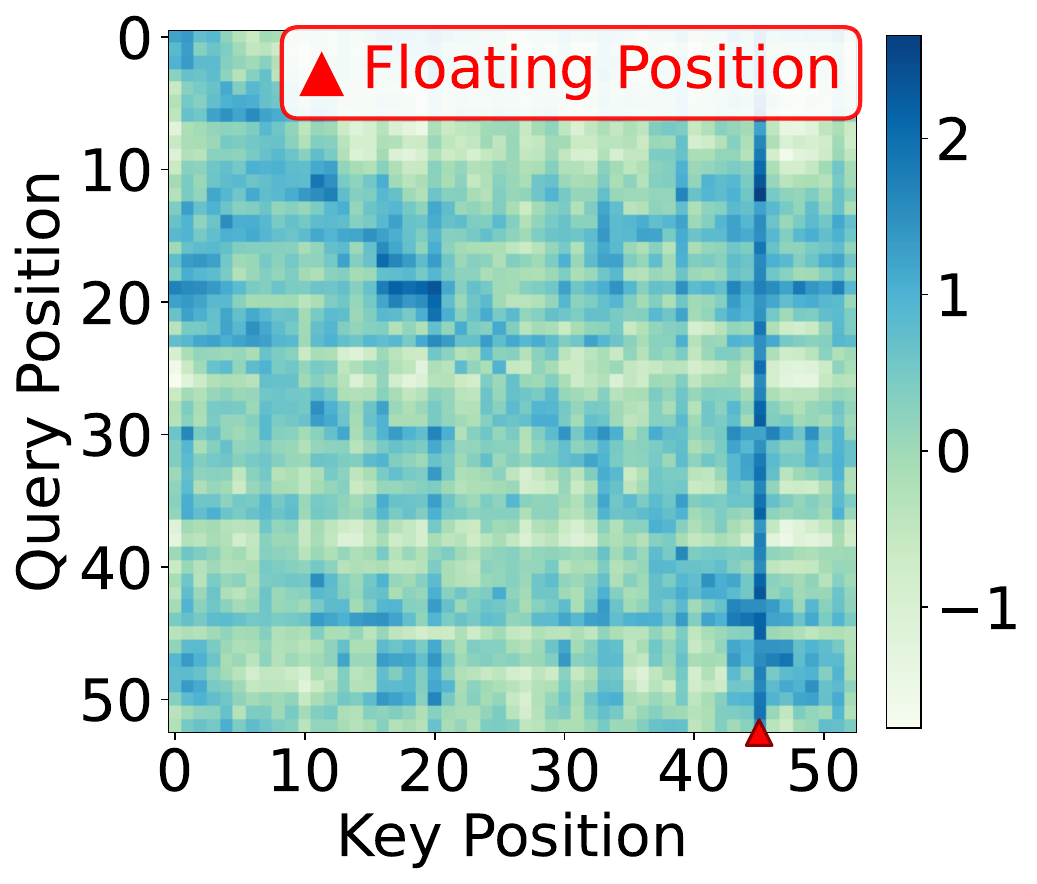}
        \caption{QK Score.}
        \label{fig:qk_decomp_l28_qk}
    \end{subfigure}\hfill
    \begin{subfigure}[t]{0.32\linewidth}
        \centering
        \includegraphics[width=\linewidth]{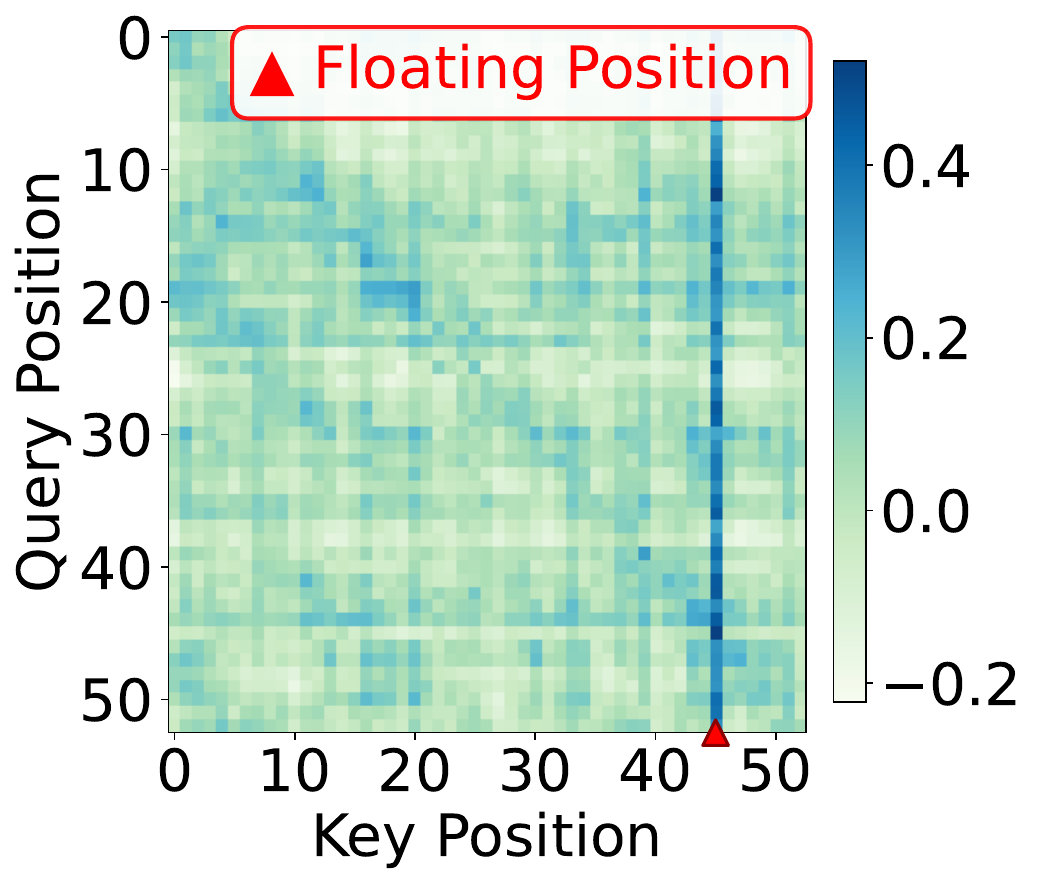}
        \caption{Angular.}
        \label{fig:qk_decomp_l28_cos}
    \end{subfigure}\hfill
    \begin{subfigure}[t]{0.32\linewidth}
        \centering
        \includegraphics[width=\linewidth]{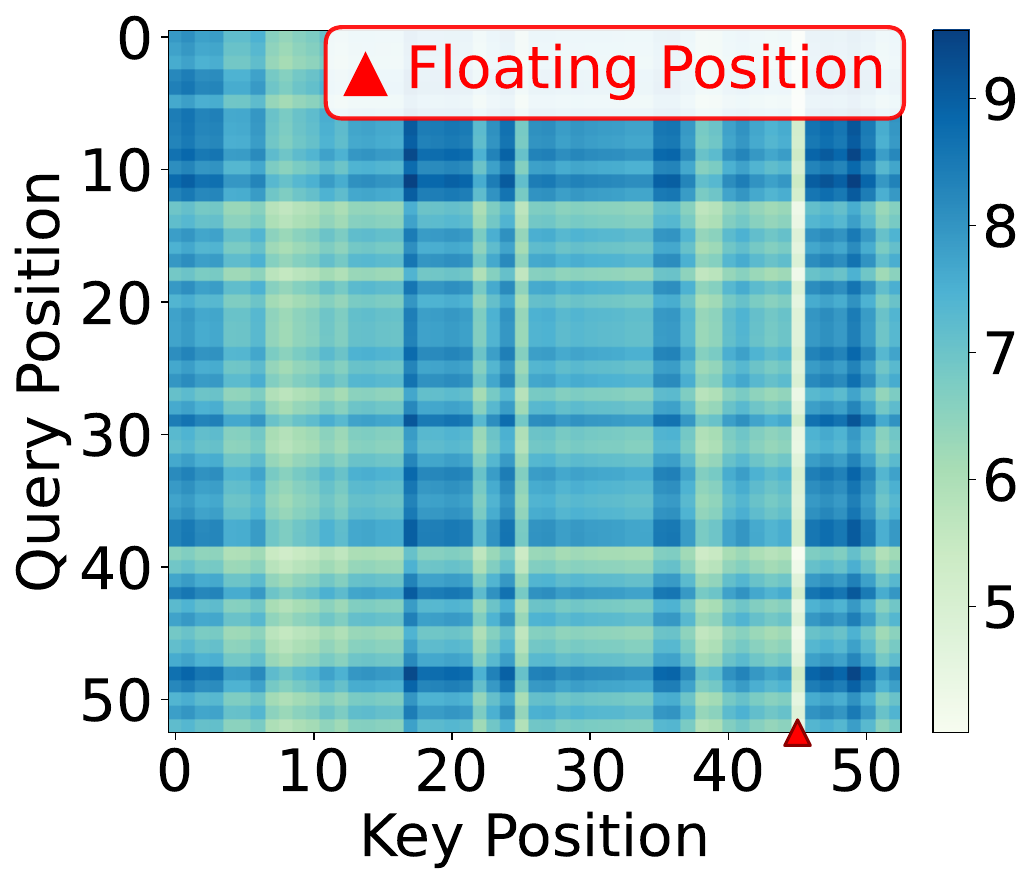}
        \caption{Norm Product.}
        \label{fig:qk_decomp_l28_norm}
    \end{subfigure}

    \vspace{2pt}
    {\small\textit{Layer 28.}}\par

    \caption{QK Geometric Decomposition across Different Layers in MDM.}
    \label{fig:llada_qk_decomposition_layers}
\end{figure}

Figure~\ref{fig:llada_qk_decomposition_layers} illustrates the QK decomposition, and heatmaps covering all layers are provided in Appendix~\ref{qk_appen}. The horizontal axis corresponds to key positions, while the vertical axis corresponds to query positions. These floating tokens, identified from attention statistics, are marked along the horizontal axis. At each depth, we jointly present the pre-softmax score map (QK Score) together with its decomposed directional alignment component (Angular) and scale component (Norm Product). This visualization enables a direct in-depth comparison to assess whether the QK advantage of floating key columns over other key columns is mainly attributable to stronger angular alignment, scale amplification, or their combined effect.

The visualization results for MDMs indicate that, in shallow layers, the QK distribution (Figure~\ref{fig:qk_decomp_l3_qk}) is relatively dispersed, with no clear contrast formed between floating and non-floating positions. This suggests that shallow-layer attention is still in an exploratory phase, not yet having established a stable structural anchoring pattern.
As depth increases, the QK contrast between floating and non-floating positions reaches its peak: in Figure~\ref{fig:qk_decomp_l11_qk}, the vertical stripes at floating positions become particularly prominent, while the QK scores in non-floating regions remain relatively low. Combined with our empirical finding in Section~\ref{structure} that the floating region is almost entirely occupied by structural tokens, this pattern suggests that attention at this stage preferentially latches onto structural anchors to stabilize denoising and information aggregation. At this stage, the QK score advantage of floating tokens is jointly driven by direction (Figure~\ref{fig:qk_decomp_l11_cos}) and scale (Figure~\ref{fig:qk_decomp_l11_norm}).
In deeper layers (roughly after layer 20), the scale term in Figure~\ref{fig:qk_decomp_l28_norm} exhibits an opposite trend: floating positions show substantially smaller scale magnitudes than non-floating positions. Consequently, the scale term no longer contributes positively to the elevated QK scores of floating columns, and the remaining advantage of the QK score can be attributed more consistently to the directional alignment term (Figure~\ref{fig:qk_decomp_l28_cos}) itself. Meanwhile, the QK scores at non-floating positions also rise noticeably, causing the overall distribution between floating and non-floating positions to become more balanced. This indicates that deep-layer attention, while maintaining focus on structural anchors, also begins to allocate more weight to tokens carrying semantic content.

Synthesizing these layer-wise patterns, we propose the following hypothesis: shallow layers are still exploring global information; as depth increases, attention relies more heavily on structural anchors and establishes a stable structural framework; deep layers then build upon this framework by progressively reallocating the attention toward tokens that carry semantic content. We refer to this transition from shallow exploration, through increasingly strong structural anchoring, to deep-layer content-driven behavior as the \emph{Shallow Structure-Aware, Deep Content-Focused} attention mechanism.

\begin{figure}[t]
  \centering
  \begin{minipage}{\columnwidth}
    \centering
    \begin{subfigure}[b]{0.48\linewidth}
      \centering
      \includegraphics[width=\linewidth]{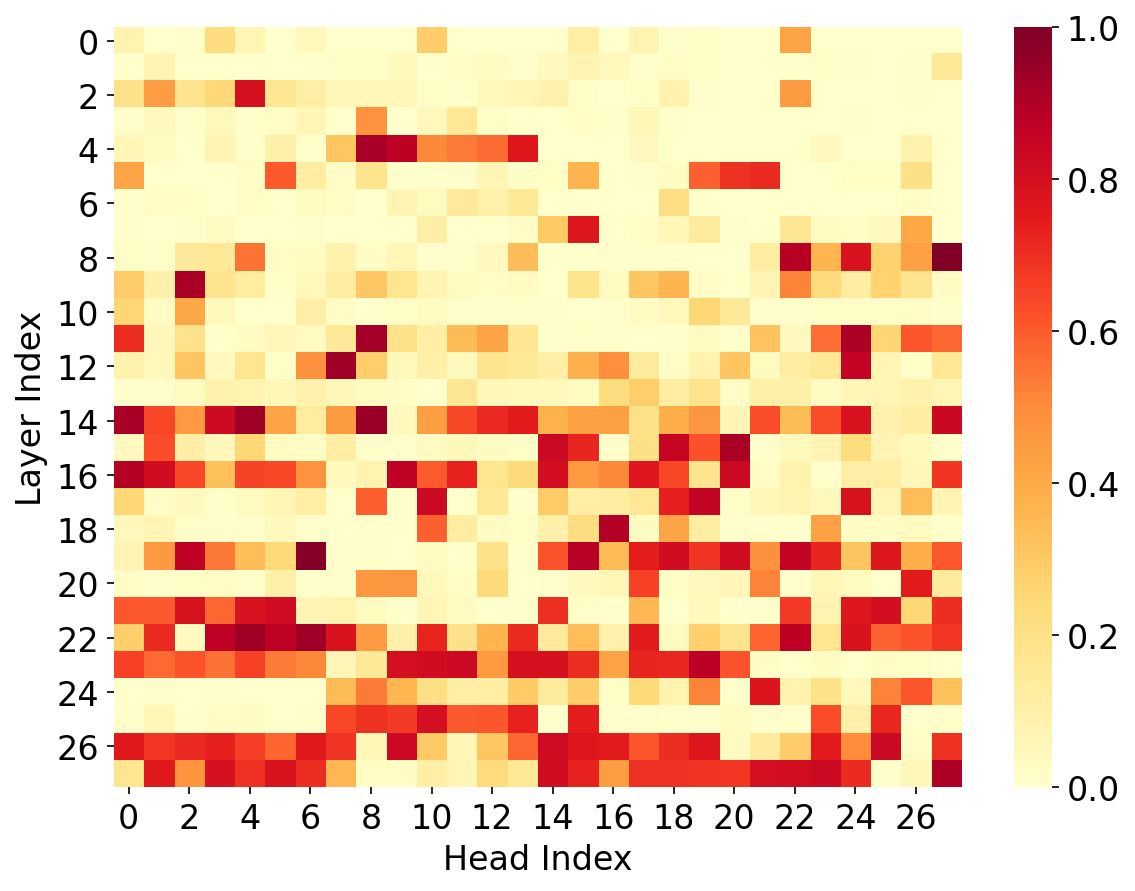}%
      \caption{Dream.}
      \label{fig:head_ar}
    \end{subfigure}%
    \hfill
    \begin{subfigure}[b]{0.48\linewidth}
      \centering
      \includegraphics[width=\linewidth]{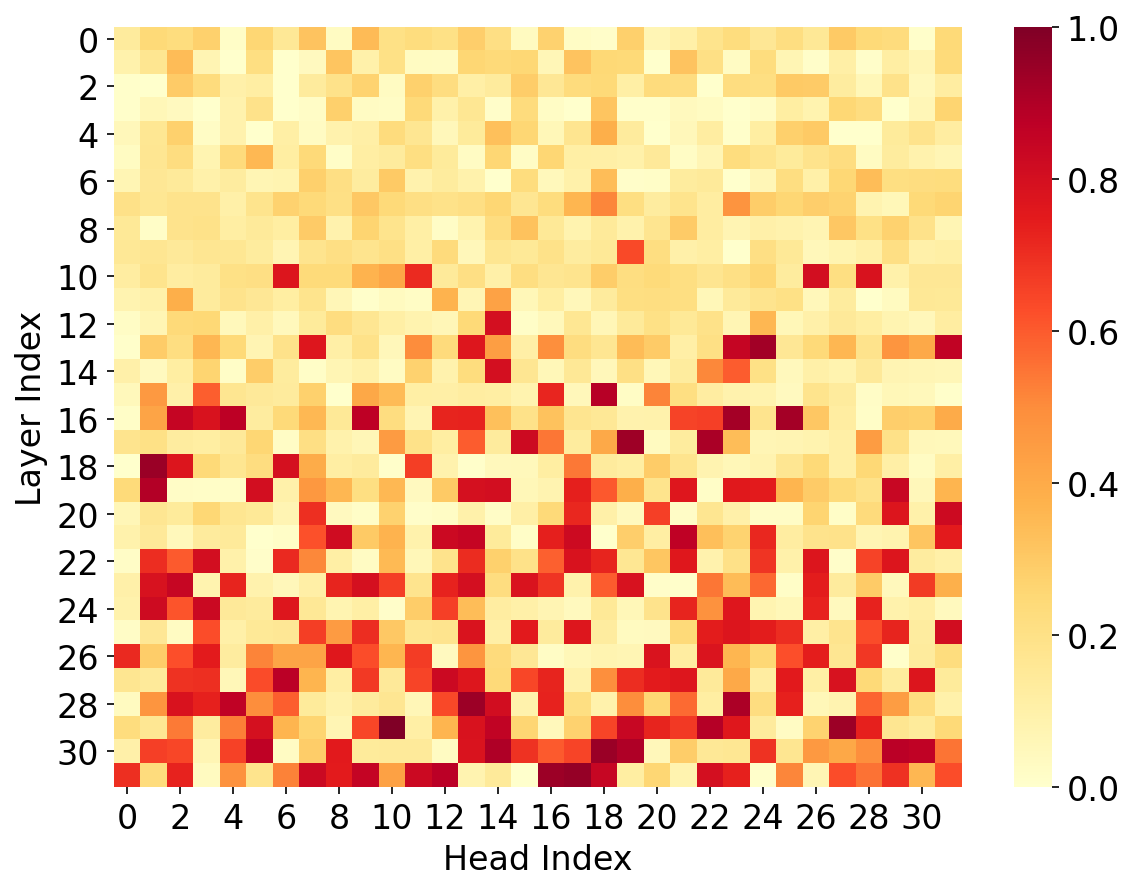}%
      \caption{Llada.}
      \label{fig:head_MDM}
    \end{subfigure}
    \caption{Retrieval Head Analysis in MDMs.}
    \label{fig:head_depthwise}
  \end{minipage}
\end{figure}

\begin{table*}[htbp]
\centering
\small
\begin{tabular}{@{}lccccccccccc@{}}
\toprule
& \multicolumn{6}{c}{\textbf{Open-Domain QA}} & \multicolumn{2}{c}{\textbf{Multi-hop QA}} & \multicolumn{2}{c}{\textbf{Slot Filling}} &  \\
\cmidrule(lr){2-7} \cmidrule(lr){8-9} \cmidrule(lr){10-11}
\textbf{Model} & \multicolumn{2}{c}{\textbf{NQ}} & \multicolumn{2}{c}{\textbf{TQA}} & \multicolumn{2}{c}{\textbf{Marco QA}} & \multicolumn{2}{c}{\textbf{HotpotQA}} & \multicolumn{2}{c}{\textbf{T-REx}} & \textbf{Avg.} \\
\cmidrule(lr){2-3} \cmidrule(lr){4-5} \cmidrule(lr){6-7} \cmidrule(lr){8-9} \cmidrule(lr){10-11}
& \textit{Acc} & \textit{F1} & \textit{Acc} & \textit{F1} & \textit{Acc} & \textit{F1} & \textit{Acc} & \textit{F1} & \textit{Acc} & \textit{F1} \\
\midrule
\multicolumn{12}{c}{\textbf{Autoregressive Models}} \\
\hdashline
Llama & 15.69 & 32.07 & 61.52 & 70.78 & 0.30 & 16.78 & 10.61 & 23.83 & 23.28 & 34.01 & 28.89 \\
Llama w/ RAG  & 35.92 & 50.63 & 74.84 & 81.15 & 0.90 & 15.15 & 20.93 & 29.80 & 24.60 & 28.61 & 36.25 \textcolor{orange}{(+7.37)} \\
Qwen & 14.95 & 27.20 & 52.58 & 59.14 & 0.20 & 11.20 & 19.05 & 27.75 & 31.20 & 37.65 & 28.09 \\
Qwen w/ RAG  & 34.16 & 50.07 & 72.61 & 80.35 & 0.50 & 16.11 & 24.50 & 34.05 & 30.02 & 34.97 & 37.74 \textcolor{orange}{(+9.64)} \\
\midrule
\multicolumn{12}{c}{\textbf{Masked Diffusion Models}} \\
\hdashline
Dream & 17.48 & 27.13 & 47.81 & 53.37 & 0.53 & 13.36 & 16.96 & 24.30 & 27.54 & 33.22 & 26.16 \\
Dream w/ RAG  & 38.66 & 53.22 & 76.39 & 82.32 & 0.90 & 19.25 & 24.23 & 34.17 & 29.46 & 35.42 & 39.39 \textcolor{red}{(+13.23)} \\
Llada & 11.77 & 21.47 & 37.11 & 42.88 & 0.70 & 11.48 & 15.18 & 22.55 & 28.92 & 33.26 & 22.52 \\
Llada w/ RAG  & \textbf{55.30} & \textbf{59.78} & \textbf{84.90} & \textbf{85.89} & \textbf{2.33} & \textbf{30.52} & \textbf{35.79} & \textbf{39.40} & \textbf{43.02} & \textbf{45.09} & \textbf{48.20} \textcolor{red}{(+25.68)} \\
% Llada2 & 16.00 & 30.45 & 46.30 & 55.66 & 0.83 & 12.72 & 14.04 & 27.23 & 28.68 & 36.25 & 26.82 \\
% Llada2 w/ RAG & 49.89 & 54.82 & 82.27 & 83.21 & 1.70 & 16.40 & 28.72 & 33.74 & 33.43 & 35.47 & 41.97 \textcolor{red}{(+15.15)} \\
\bottomrule
\end{tabular}
\caption{Overall Performance of ARMs and MDMs. The \textbf{best} results are highlighted.}
\label{tab:model_comparison}
\end{table*}

\textbf{Retrieval Head Analyses.}
To further verify the above hypothesis, we conduct an analysis following~\citet{wu2024retrieval} on retrieval-specialized heads. 

Specifically, we assign each attention head a retrieval score that quantifies how frequently the head allocates top-$k$ attention to the ground-truth needle tokens during final answer generation. A higher score indicates that the head effectively routes information from relevant context tokens into the output, whereas a lower score suggests that the head rarely contributes to context-following behavior.
The resulting retrieval-score heatmap (Figure~\ref{fig:head_depthwise}) reveals that, in MDMs, high-scoring heads are predominantly concentrated in the middle and deeper layers.
This pattern suggests that, with increasing depth, MDMs progressively allocate greater attention capacity to content-sensitive retrieval heads that track context-bearing tokens. Such a depth-wise transition from structurally oriented floating behavior to content-centric retrieval behavior is exactly in line with the \emph{Shallow Structure-Aware, Deep Content-Focused} attention mechanism hypothesis.

\section{Attention Floating Improves Robustness in Learning from Context}
\label{sec5}
% In this section, we investigate how the attention floating mechanism impacts the model's capability to learn from context. We begin by evaluating the overall performance across various Question-Answering (QA) tasks, examining how effectively different model architectures incorporate external knowledge to support generation. We then characterize the information aggregation process by visualizing the attention flow in multi-document settings. Finally, we conduct a focused analysis on robustness, investigating how MDMs perform under challenging conditions such as long-range dependencies, scattered evidence, and varying levels of contextual noise.

In this section, we transition from analyzing internal mechanisms to empirically evaluating the model's capability to learn from context.
We begin in Section~\ref{5.1} by evaluating overall performance on a range of knowledge-intensive tasks, where we find that MDMs benefit more substantially from retrieved context than ARMs. Then, to better understand the sources of this performance disparity, 
Section~\ref{5.2} shows that \emph{attention floating} contributes to robust learning from context through a set of stress tests. Finally, Section~\ref{5.3} investigates the information aggregation process through region-level attention flow, offering a mechanistic explanation for the empirically observed gains.

\subsection{Performance of ARMs and MDMs in Learning Knowledge from Contexts}
\label{5.1}
This section examines the overall performance of autoregressive models (ARMs) and masked diffusion models (MDMs) across knowledge-intensive tasks, with and without retrieved context. Detailed dataset statistics are provided in Appendix~\ref{app:data-stats}. 

As shown in Table~\ref{tab:model_comparison}, autoregressive baselines achieve slightly higher average scores than MDMs in the close-book QA setting, which is consistent with their stronger parametric capacity. However, once we incorporate query-retrieved passages as contextual input, MDMs not only close the performance gap with ARMs, but also outperform all ARM w/ RAG baselines across all QA scenarios.
MDMs w/ RAG achieve over 19.5\% average improvement compared to their corresponding baseline models, which is more than twice the gain observed for ARMs when augmented with retrieval (ARMs w/ RAG obtain 8.5\% improvements). These results indicate that, although MDMs start from a slightly weaker parametric baseline, they are substantially more effective at transforming retrieved evidence into end-task performance gains. In the following experiments, we further investigate how \textit{attention floating} enables a more retrieval-sensitive and context-driven utilization of external knowledge.

%为了进一步探究外部上下文的引入对模型性能的影响，我们对QA数据集的topk检索内容逐步引入噪音，观察其性能变化，并改变gold evidence在序列中的位置，观察表现的波动。
% \subsection{Robustness to Contextual Variations}
\subsection{Effectiveness of Attention Floating with Contextual Stress Testing}
\label{5.2}
To further examine how the \emph{attention floating} supports robustness under context variations, we conduct a systematic evaluation along three key dimensions: (i) contextual noise interference, (ii) position perturbation, and (iii) evidence integration. 
% Specifically, we progressively inject increasing amounts of distracting noise into the top retrieved content to assess performance degradation, and vary the position of the gold document within the input sequence to analyze positional sensitivity and model robustness.

\begin{figure}[t]
    \centering
    \includegraphics[width=\linewidth]{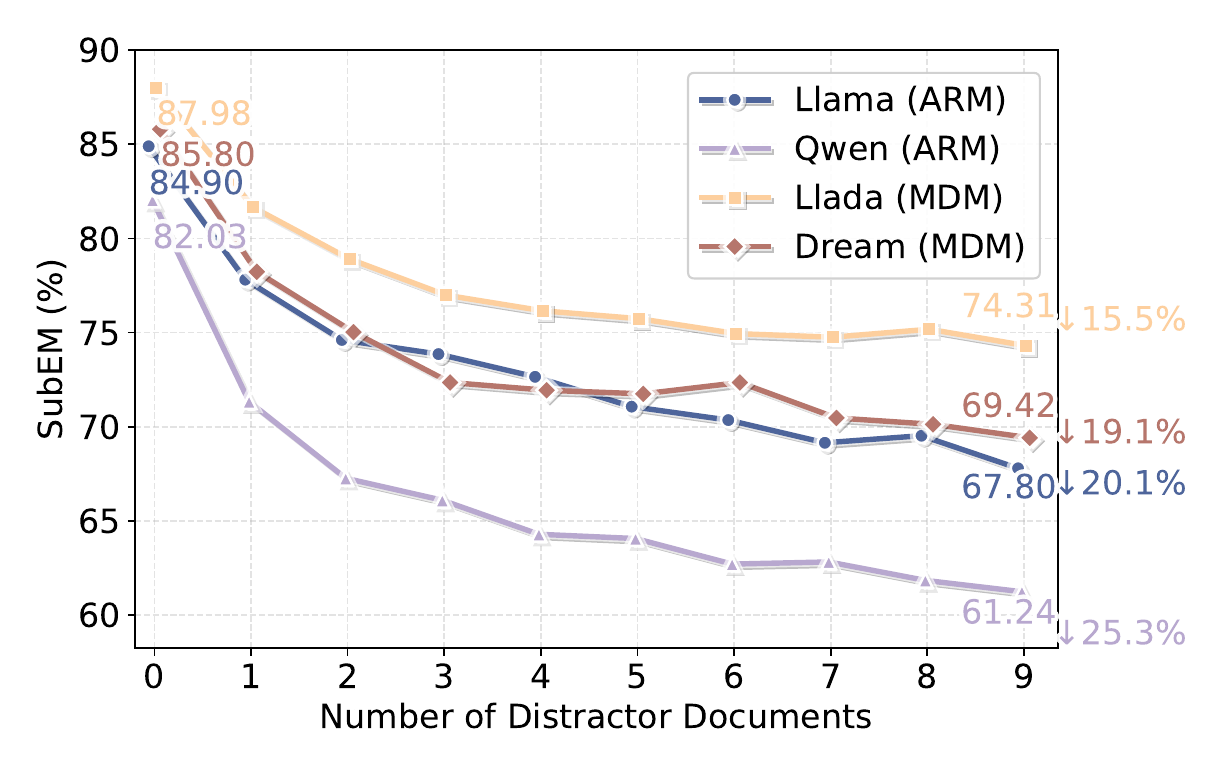}
    \caption{Performance with Varying Numbers of Distractor Documents.}
    \label{fig:variant}
\end{figure}

\textbf{Contextual Noise Interference.}
To evaluate model robustness under different signal-to-noise conditions, we adopt an experimental setup following prior work~\cite{hsieh2024ruler}: we fix the context to contain exactly one gold document, gradually increase the number of unrelated distractor documents, and observe how model performance changes.
As shown in Figure~\ref{fig:variant}, as the number of distractor documents increases, the performance of the ARMs exhibits a gradual degradation, which may be attributed to the autoregressive generation architecture. In contrast, the MDM demonstrates substantially stronger noise resilience: even when the number of distractor documents is significantly increased, its accuracy degrades more mildly. This further indicates that MDMs are better at alleviating the impact of irrelevant documents, supporting the view that the bidirectional attention mechanism of MDMs provides a form of global denoising.

\begin{figure}[t]
  \centering
  \begin{minipage}{\columnwidth}
    \centering
    \begin{subfigure}[b]{0.48\linewidth}
      \centering
      \includegraphics[width=\linewidth]{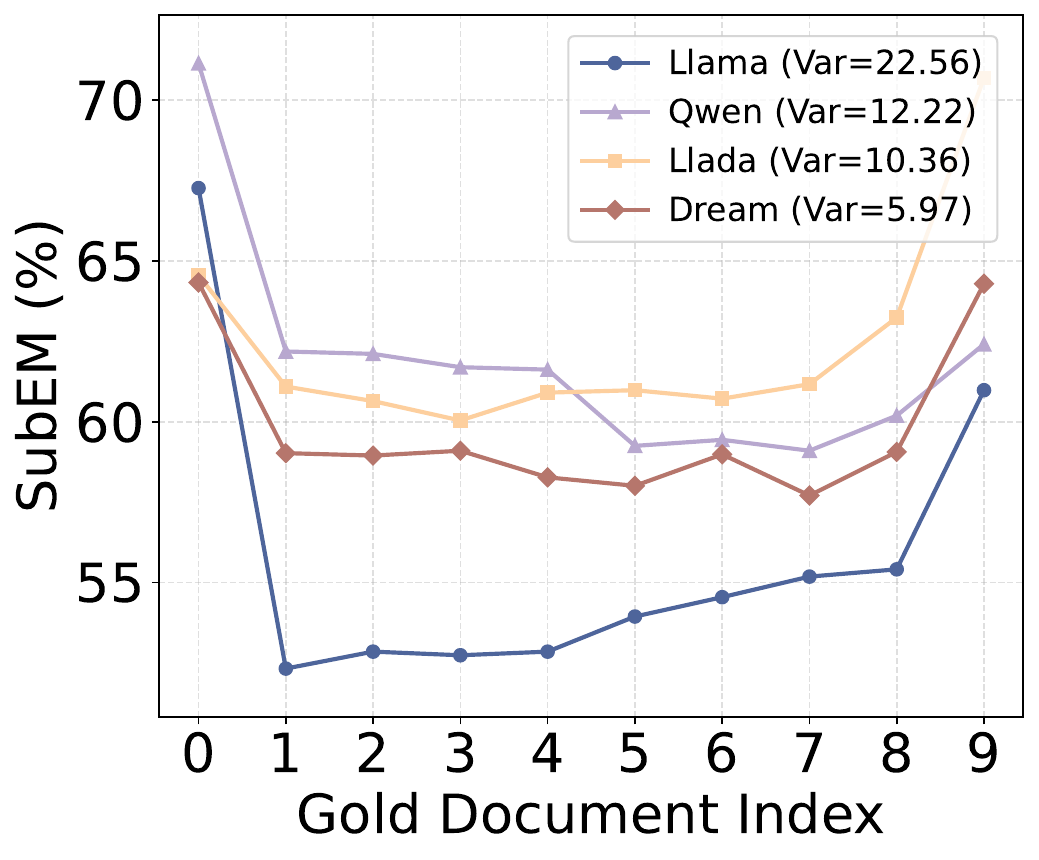}%
      \caption{Position Shift. }
      \label{fig:position}
    \end{subfigure}%
    \hfill
    \begin{subfigure}[b]{0.48\linewidth}
      \centering
      \includegraphics[width=\linewidth]{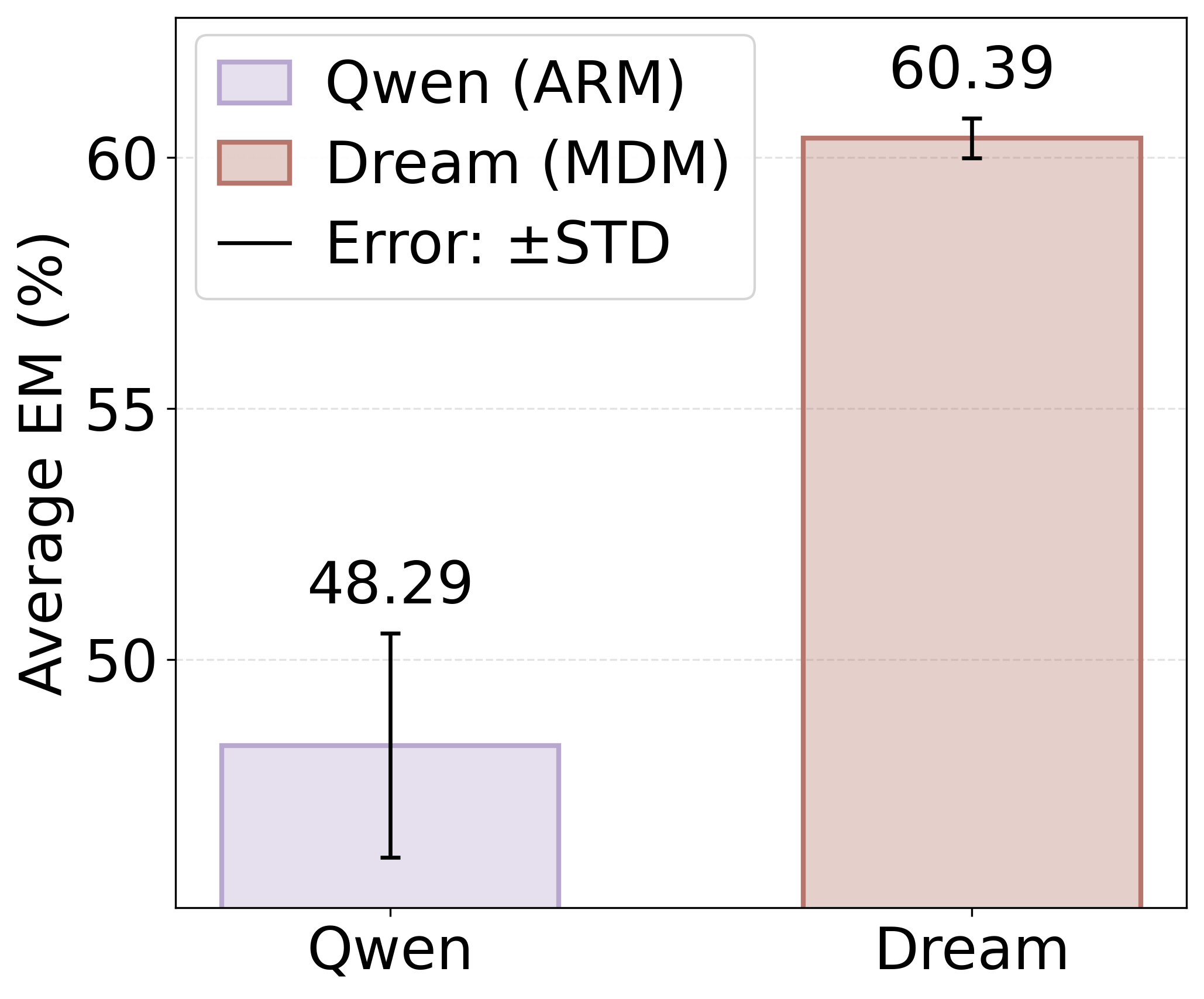}%
      \caption{Evidence Integration.}
      \label{fig:multi-hop}
    \end{subfigure}
    \caption{Performance under (a) Position Perturbation and (b) Evidence Integration Scenarios.}
    \label{fig:rag-long-multihop}
  \end{minipage}
\end{figure}

\textbf{Position Perturbation.}
Beyond the impact of contextual noise, the position of key information within a long context also affects model performance~\cite{liu2024lost}. Specifically, we evaluate a multi-document QA task by systematically varying the position of the gold document among distractors to examine the model's sensitivity to evidence location.
As illustrated in Figure~\ref{fig:position}, ARMs exhibit a characteristic U-shaped performance curve, with accuracy peaking when the gold evidence is located near the boundaries and deteriorating when it appears in the middle. In contrast, MDMs show substantially smaller performance variance across different positions, suggesting that they are less sensitive to the location of key information. This positional robustness can be attributed to the \emph{attention floating} mechanism: Unlike ARMs, which exhibit rigid sinking around <BOS> and a strong recency bias, MDMs can actively reorganize their attention distribution.

\textbf{Evidence Integration.}
We further investigate whether the architecture of MDMs leads to more robust behavior in complex scenarios, 
particularly in multi-hop reasoning tasks where the reasoning outcome requires the integration of information from scattered evidence. Specifically, while keeping the content unchanged, we systematically perturb the distribution of these evidence documents within the input context.
As shown in Figure \ref{fig:multi-hop}, the ARM exhibits pronounced sensitivity to different evidence distribution, as evidenced by achieving a higher variance score than MDMs. In contrast, the MDM achieves superior average performance while maintaining remarkable stability. The minimal variance indicates that the MDM is largely insensitive to the distribution of gold documents, demonstrating its robust capability to effectively integrate evidence from retrieved documents.
\subsection{Attention Floating Analysis via Region-Level Attention Flow}
\label{5.3}
To better understand the underlying mechanisms that lead to the superior performance of MDMs across different scenarios, we analyze their internal information flow through the lens of attention mechanisms. This perspective enables a fine-grained examination of how models dynamically allocate attention across different input regions.

\begin{figure}[t]
    \centering
    % ---- 第一行 ----
    \begin{subfigure}[b]{0.48\linewidth}
        \centering
        \includegraphics[width=\linewidth]{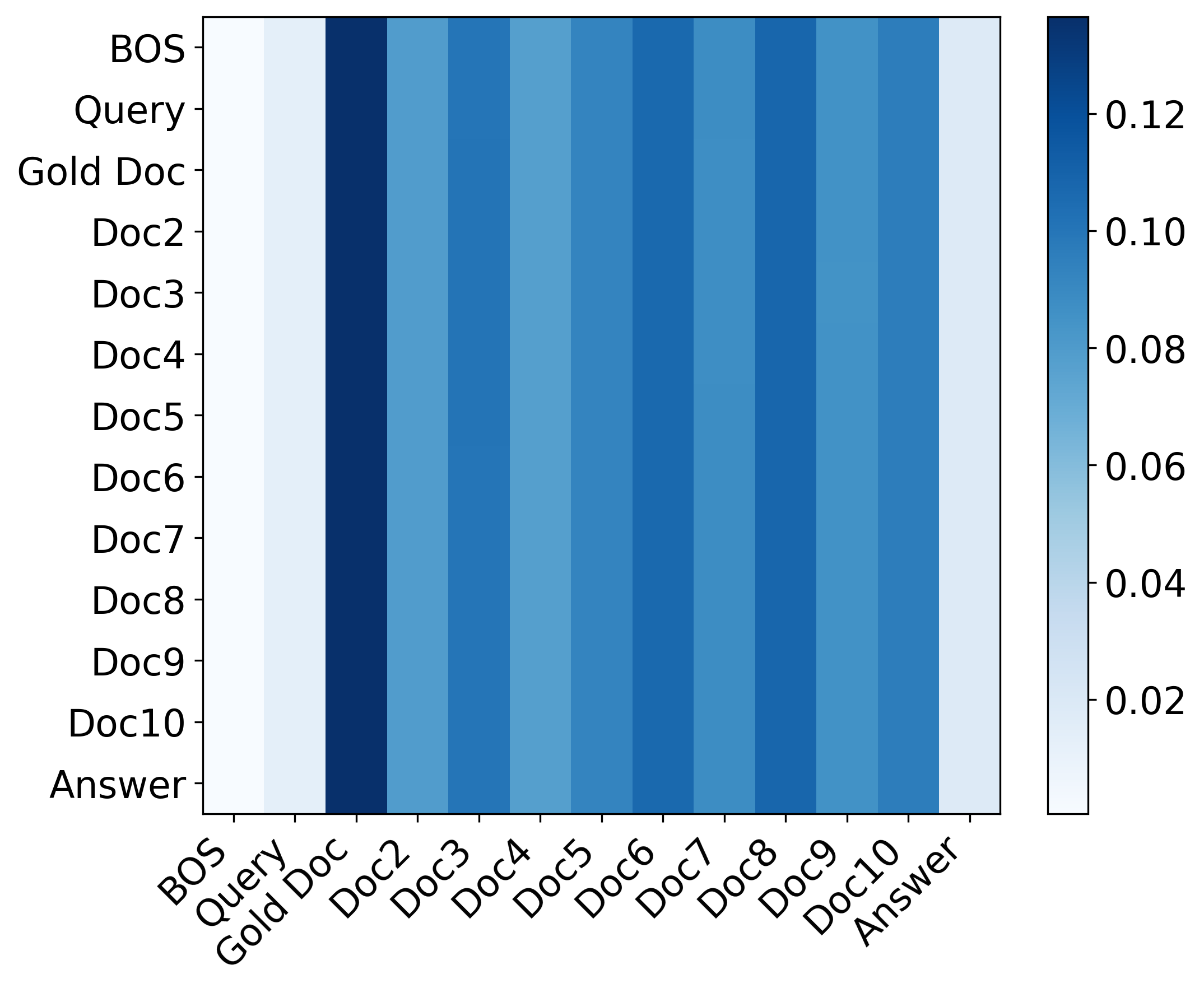}
        \caption{Llada with the Gold Doc at Position 1.}
        \label{fig:rollout-gold0}
    \end{subfigure}\hfill
    \begin{subfigure}[b]{0.48\linewidth}
        \centering
        \includegraphics[width=\linewidth]{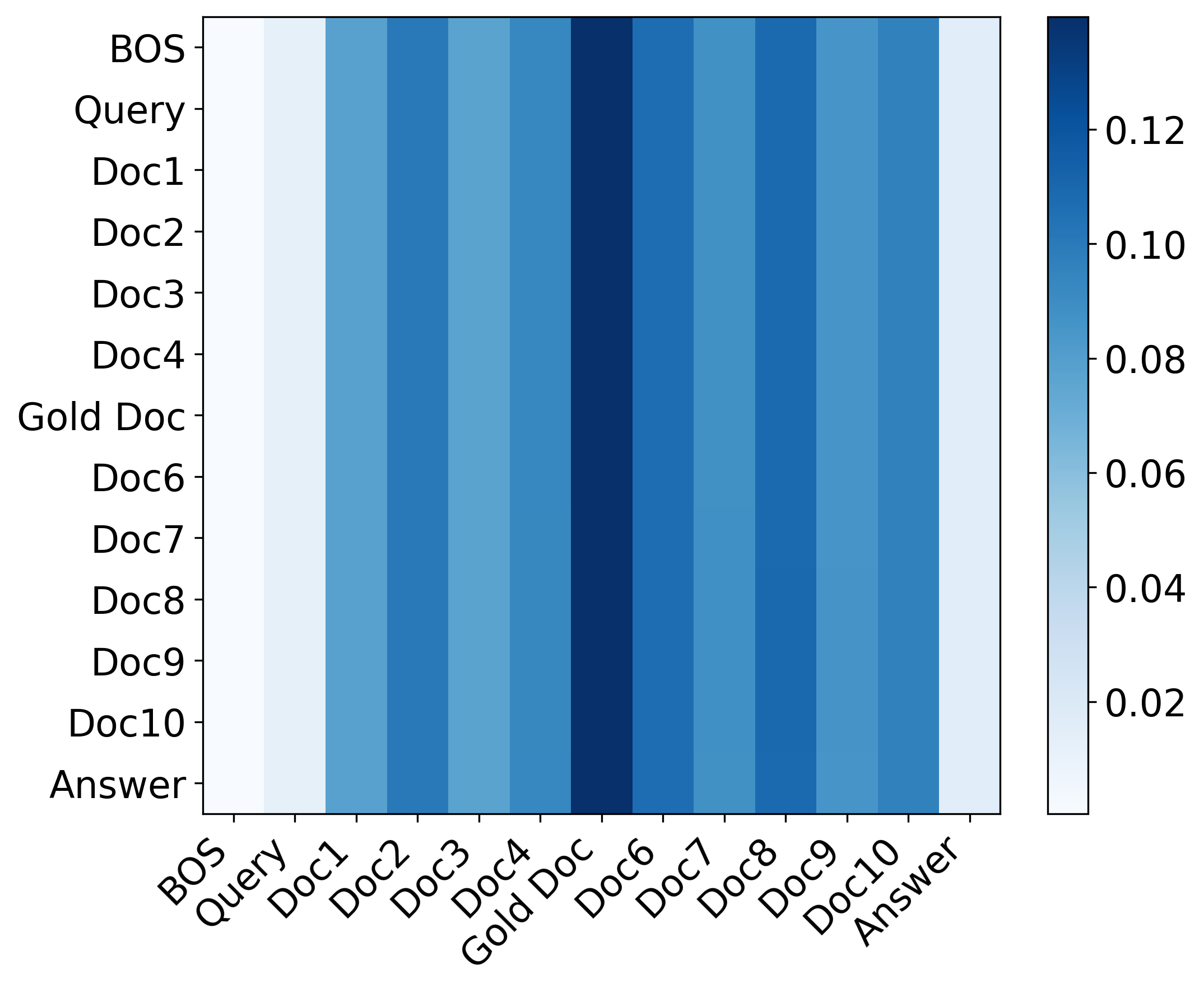}
        \caption{Llada with the Gold Doc at Position 5.}
        \label{fig:rollout-gold4}
    \end{subfigure}

    \vspace{0.5em}
    % ---- 第二行 ----
    \begin{subfigure}[b]{0.48\linewidth}
        \centering
        \includegraphics[width=\linewidth]{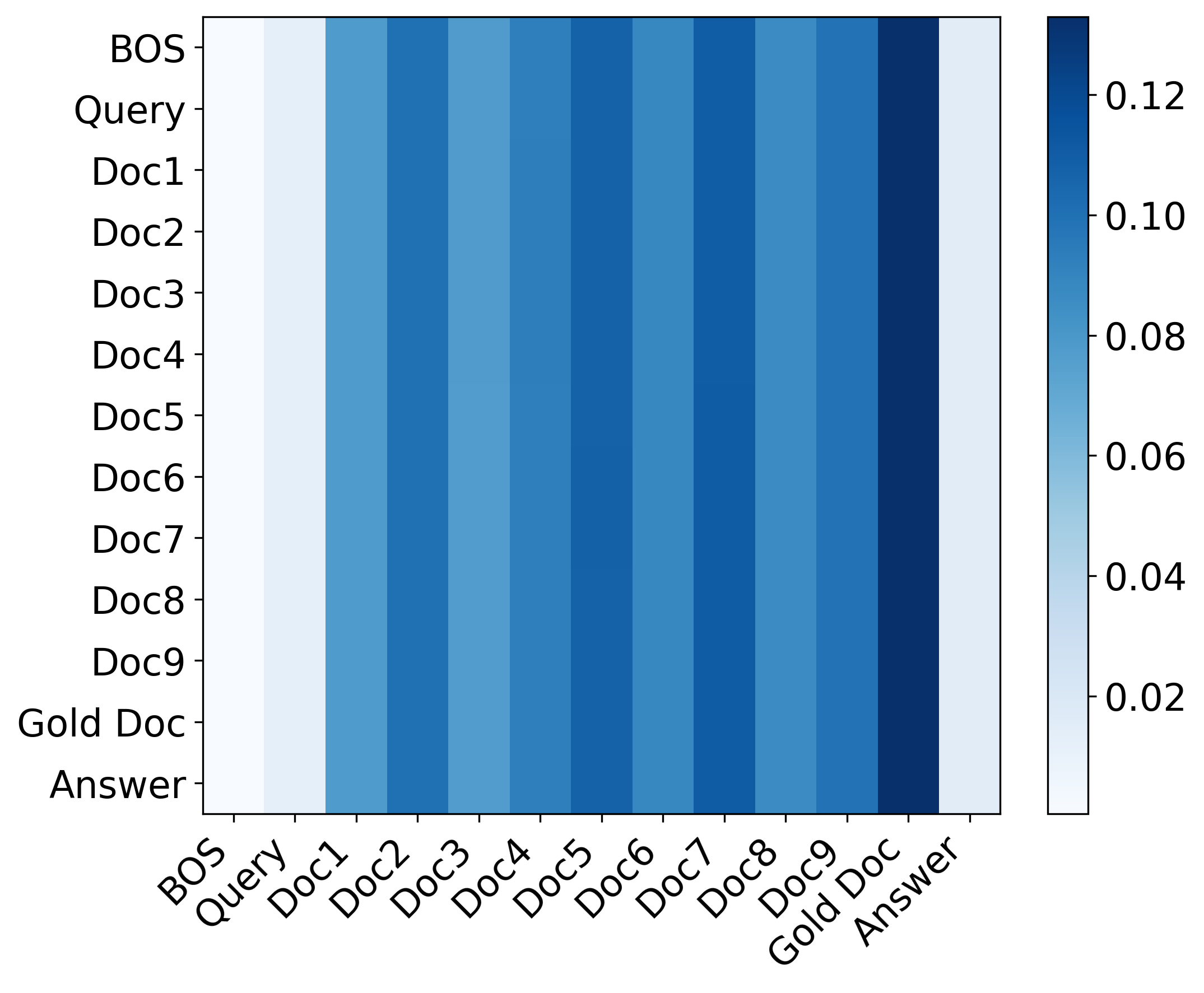}
        \caption{Llada with the Gold Doc at Position 10.}
        \label{fig:rollout-gold9}
    \end{subfigure}\hfill
    \begin{subfigure}[b]{0.48\linewidth}
        \centering
        \includegraphics[width=\linewidth]{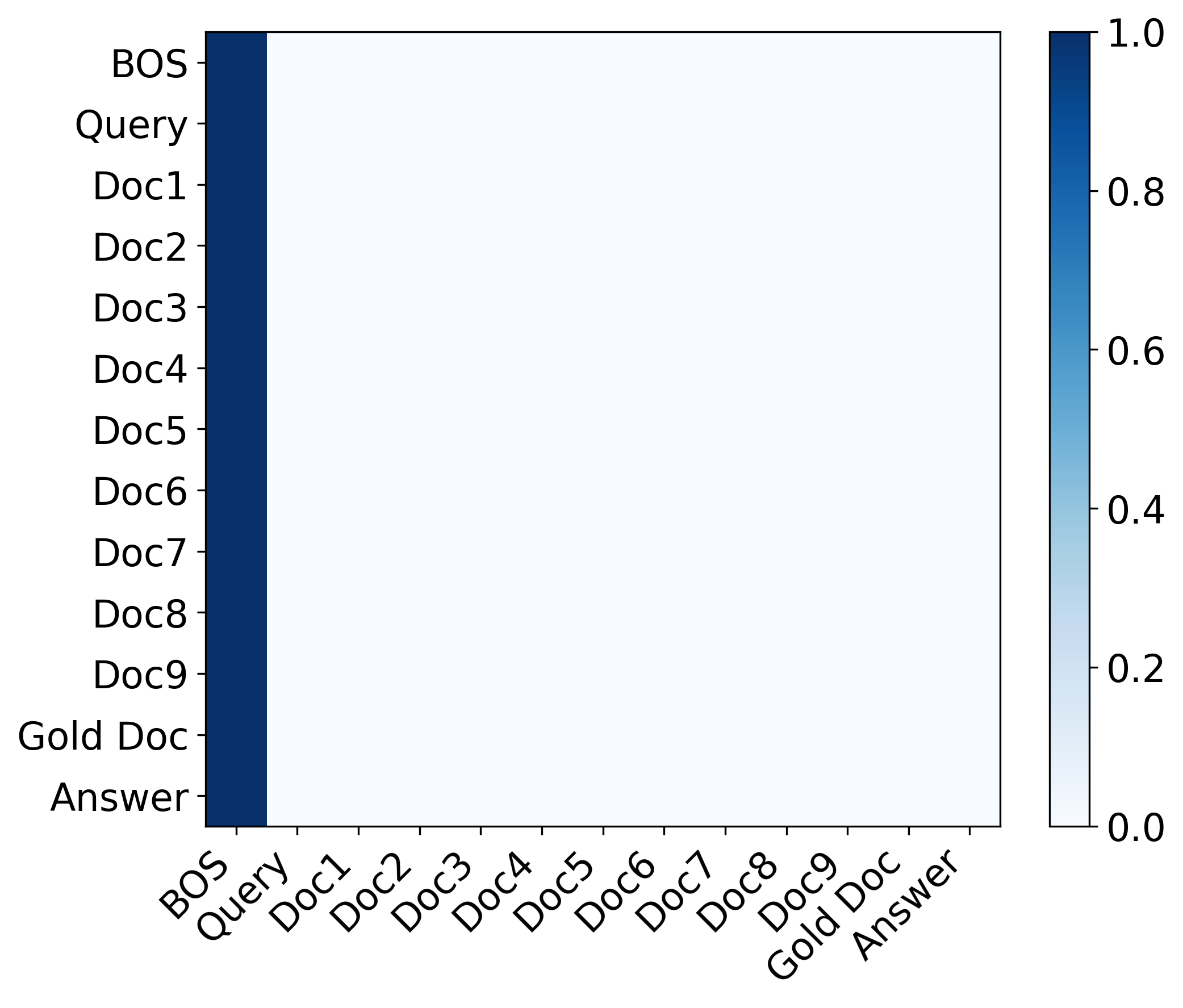}
        \caption{Llama with Gold Doc at Positions 1, 5, 10.}
        \label{fig:rollout-llama9}
    \end{subfigure}

    \caption{
    Region-Level Attention Flow of Llada (a–c) and Llama (d) with Gold Doc at Different Positions.}
    \label{fig:rollout-four}
\end{figure}

To characterize the information flow across all layers, we adopt an attention-flow-style region-level influence matrix~\cite{abnar2020quantifying}. Concretely, we aggregate head-averaged attention across layers into a position-level influence matrix, and subsequently group contiguous positions into coarse-grained regions corresponding to <BOS>, Query, Doc1–Doc10, and Answer. The formal definition of the attention flow procedure is provided in Appendix~\ref{app:rollout-multi-doc}. As illustrated in Figure~\ref{fig:rollout-four} and Figure~\ref{fig:rollout-four-appendix}, we observe a clear qualitative distinction between ARMs and MDMs. For ARMs, the dominant peak of the region-level flow remains tightly concentrated around <BOS>, exhibiting minimal sensitivity to changes in evidence location. In contrast, for MDMs, relocating the gold document results in a corresponding shift of the high-intensity band in the region-level flow toward the ground-truth document segments. These results indicate that the dominant attention flow in ARMs behaves as a rigid sink in <BOS> and is insensitive to the position of the gold document. In comparison, MDMs actively reorganize attention distributions to capture semantically relevant information, thus mitigating the influence of distracting documents.

% We refer to this phenomenon as the semantic alignment property of
% Attention Floating: floating tokens in MDMs do not form a hard sink at a fixed
% absolute index, but instead tend to align with genuinely useful semantic
% evidence during denoising, adapting their effective anchors to the positions of
% informative documents and answer segments.

\section{Related Work}
%从AR到DLLM
Autoregressive models (ARMs) remain the dominant paradigm but suffer from high latency and positional bias due to their sequential, causal mechanism. To alleviate these limitations, a growing body of studies~\cite{hersche2025softmaskeddiffusionlanguagemodels, liu2025tidarthinkdiffusiontalk, ni2025trainingoptimallargediffusion, shao2025diffusethinkingexploringdiffusion} on diffusion language models (DLMs) have explored replacing sequential autoregressive sampling with parallel generation via denoising, aiming to retain strong generation quality while improving parallelism and efficiency.
% DLLM -> MDM
Within the landscape of DLMs, recent research has increasingly converged on masked diffusion models (MDMs)~\cite{nie2025large, ye2025dream}, which operate natively over the discrete token vocabulary. Unlike approaches that require mapping to continuous latent spaces, MDMs typically initialize from sequences filled with special mask symbols. They define a forward process that injects masking noise and a reverse process that predicts tokens at masked positions in parallel to reconstruct the text. While early MDMs~\cite{he2023diffusionbert, zheng2023reparameterized} were relatively small in parameter and training scale, leaving a noticeable performance gap compared to ARMs, recent scaled-up MDMs~\cite{wu2025fast, liu2025longllada} have successfully narrowed this gap on general language modeling and downstream benchmarks, demonstrating their potential as a competitive alternative.

%模型机理分析的相关工作
In recent years, research on the mechanistic understanding and interpretability of large models has progressed rapidly. In ARMs, prior work has developed circuit-style analysis frameworks to elucidate decomposable attention heads and compositional mechanisms, and has demonstrated that induction heads can mechanistically account for in-context learning~\cite{elhage2021mathematical, olsson2022context}. Studies on long-context behavior have further established clear empirical baselines for positional sensitivity~\cite{liu2024lost}. In parallel, the phenomenon of attention disproportionately concentrating on early sequence positions, known as the attention sink, has been systematically analyzed~\cite{xiao2024efficient, gu2024attention}. Similar mechanistic analyses have been extended to DLMs, including investigations that dissect internal features and sources of bias, as well as work that traces the evolution of interpretable concepts along the denoising trajectory~\cite{shi2025dissecting, tinaz2025emergence}. More recently, attention-sink analyses have also been applied to DLMs, examining how sink behavior interacts with the denoising process~\cite{rulli2025attention}. Collectively, these lines of work provide both the methodological foundation and comparative reference points for our attention-mechanism analysis in MDMs.

\section{Conclusion}
% This paper investigates masked diffusion models (MDMs) from the perspective of their attention behavior. We first identify and formalize the phenomenon of attention floating, showing that MDMs exhibit weaker, mobile and structural concentration of attention. 
% We further study how attention evolves across depth, providing a more fine-grained view of information flow inside DLLMs. Finally, by examining long-dependency and multi-hop scenarios, we demonstrate that MDMs improve robustness compared to autoregressive baselines.
% We hope these findings can inform the design of future diffusion-based language models and retrieval-augmented systems, especially in challenging long-context and multi-hop reasoning settings.

This paper investigates MDMs from the perspective of their attention behavior. We first identify and formalize a phenomenon termed \textit{attention floating}, demonstrating that MDMs exhibit weaker, more mobile, and structurally less concentrated attention patterns. Our analyses reveal a Shallow Structure–Aware, Deep Content–Focused attention mechanism.
We then empirically evaluate the MDM's ability to learn from context and conduct a series of experiments to show the crucial role of the attention floating mechanism. These findings deepen the understanding of the internal working mechanisms of MDMs.
%Finally, our attention flow analysis provides a mechanistic explanation for these gains.

\section{Limitations}
This work has limitations that also point to promising future directions. First, our analysis mainly focuses on one category of Diffusion Language Models (DLLMs)--Masked Diffusion Models (MDMs). While this focus allows for a controlled and in-depth investigation of attention floating under masked denoising dynamics, we do not explicitly evaluate other DLLM variants with different diffusion or denoising mechanisms. Future work could extend this analysis framework to a broader family of DLLMs and systematically compare how different architectures and denoising strategies affect floating patterns and retrieval gains. Second, our contributions are primarily centered on mechanism analysis, and we have not yet proposed training or inference methods that can directly improve performance based on these findings; future work may explore turning floating signals into learnable regularizers or developing denoising-style context modeling for multi-document evidence integration, thereby closing the loop from mechanism to method. 

\bibliography{acl-style-files-master/custom}

\clearpage
\newpage

\appendix

\section{Appendix}

\subsection{License}
This section summarizes the licenses (or usage terms) of the datasets used in our experiments.

All datasets are used under their respective licenses and agreements, which permit academic research use:
Natural Questions (CC BY-SA 3.0 License); TriviaQA and 2WikiMQA (Apache 2.0 License); HotpotQA (CC BY-SA 4.0 License);
T-REx (CC BY-SA 4.0 License); GSM8K (MIT License), and MS MARCO QA, which is provided under the MS MARCO Terms and Conditions
for non-commercial research purposes.

\subsection{Additional Experimental Details}

\paragraph{Dataset Statistics for Knowledge-Intensive Tasks.}
\label{app:data-stats}

We evaluate all models on a suite of retrieval-augmented knowledge-intensive benchmarks that follow the evaluation configuration of \citet{lirag}. The datasets span open-domain QA, multi-hop reasoning, and slot-filling style questions. For each dataset, we use the same retrieval corpus and context construction strategy as in \citet{lirag}, and report results on evaluation-only splits. Table~\ref{tab:data-stats} summarizes the basic statistics of the datasets used in our experiments.

\begin{figure*}[t]
    \centering
    \includegraphics[width=\linewidth]{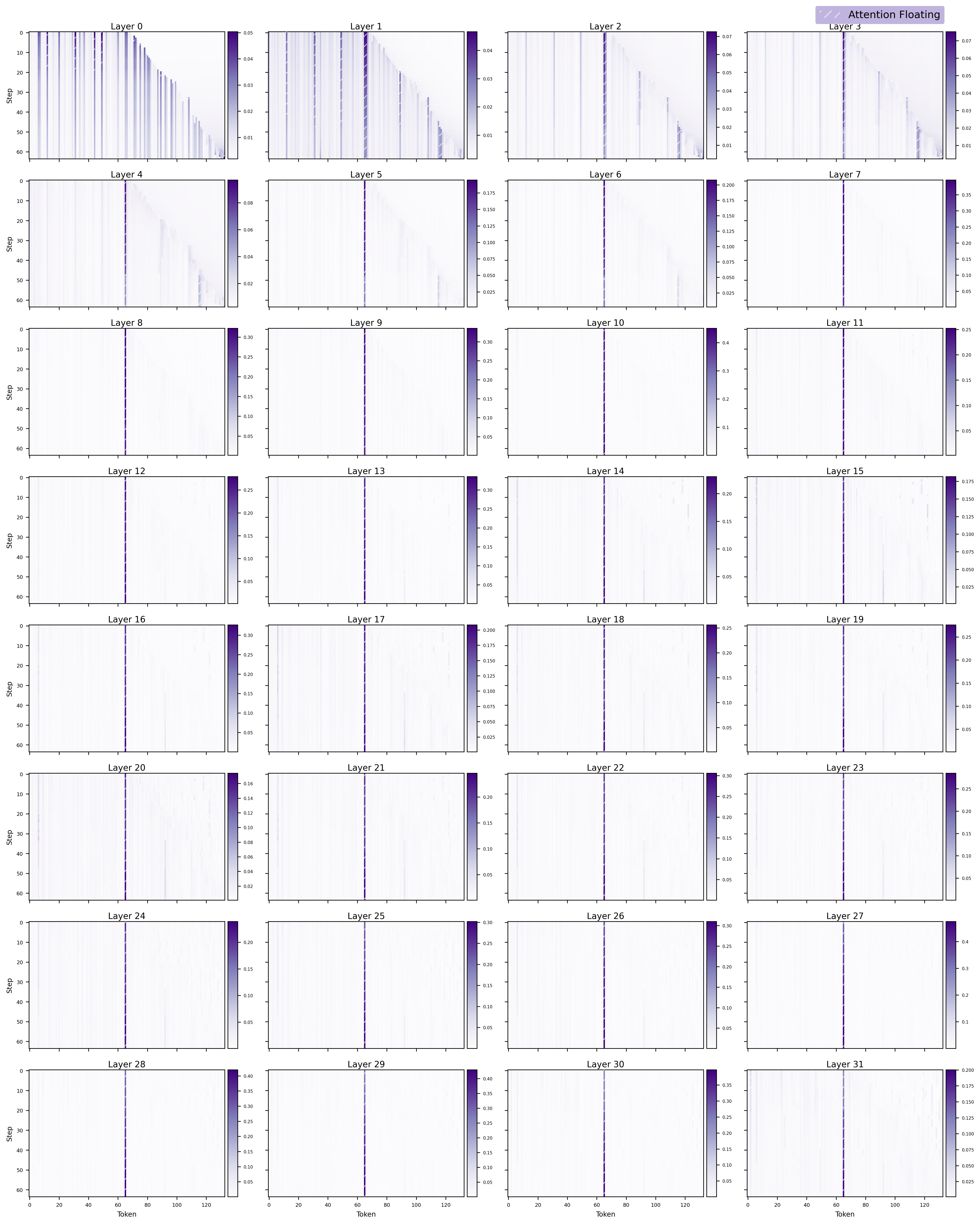}
    \caption{Positional Drift of Attention Floating in Llada.}
    \label{fig:temporal_all}
\end{figure*}

\begin{figure*}[t]
    \centering
    \includegraphics[width=\linewidth]{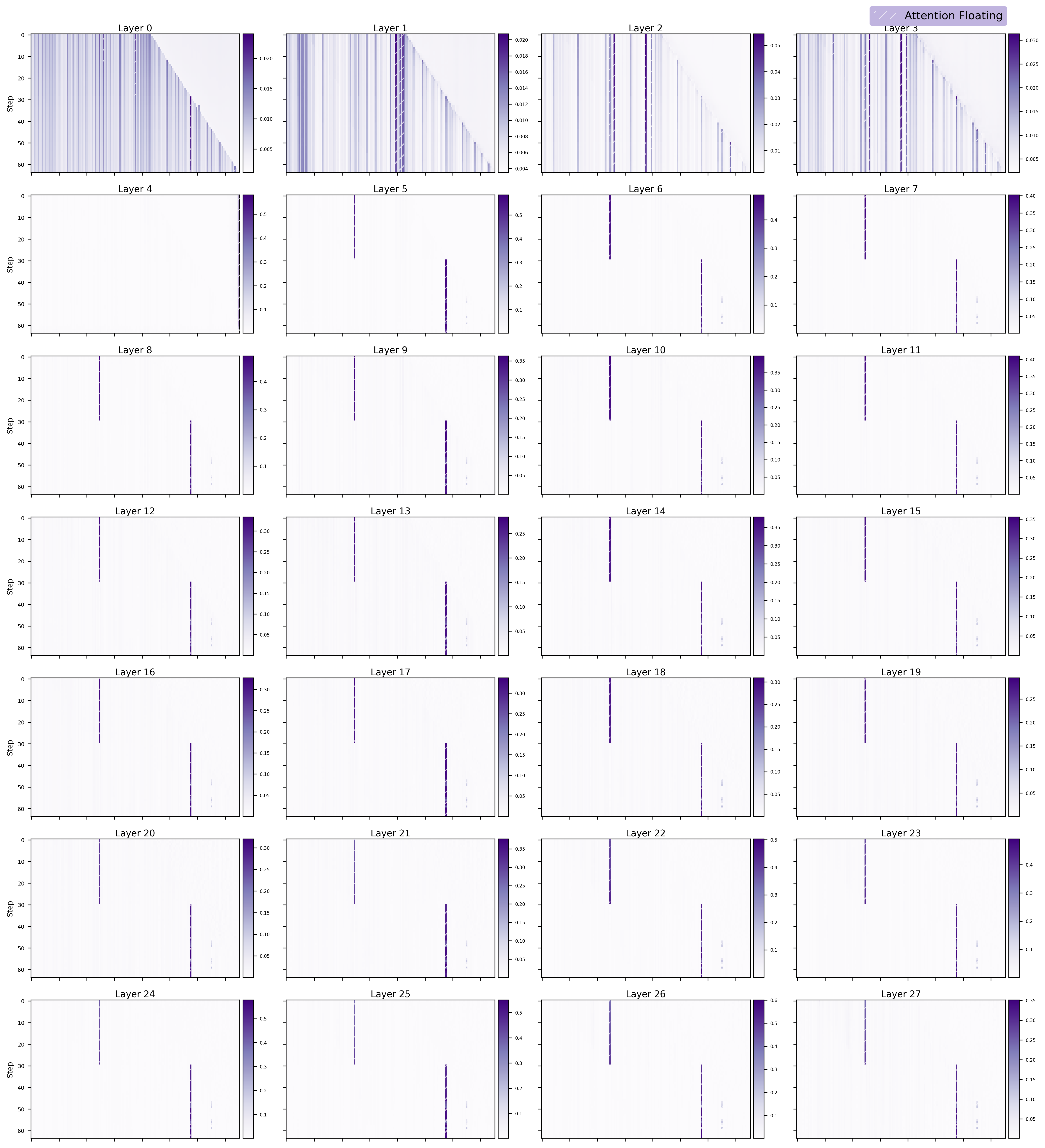}
    \caption{Positional Drift of Attention Floating in Dream.}
    \label{fig:temporal_dream}
\end{figure*}

\paragraph{Additional Visualizations of Positional Drift.}
\label{drift}
We include additional visualizations of positional drift in attention floating for Llada and Dream across denoising steps. Figure~\ref{fig:temporal_all} and Figure~\ref{fig:temporal_dream} provide qualitative support for the layer-dependent drift patterns discussed in the main text.

\paragraph{High-Frequency Floating Token Statistics.}
\label{High-Frequency}
Table~\ref{tab:floating_tokens_freq} reports the most frequent floating tokens ranked by their overall proportion. The statistics show that floating tokens are dominated by high-frequency structural symbols (e.g., newline, end-of-text, mask tokens, and punctuation), while lexical content tokens appear much less frequently, supporting our structural-controller interpretation.

\begin{figure*}[t]
    \centering
    \includegraphics[height=0.90\textheight, keepaspectratio]{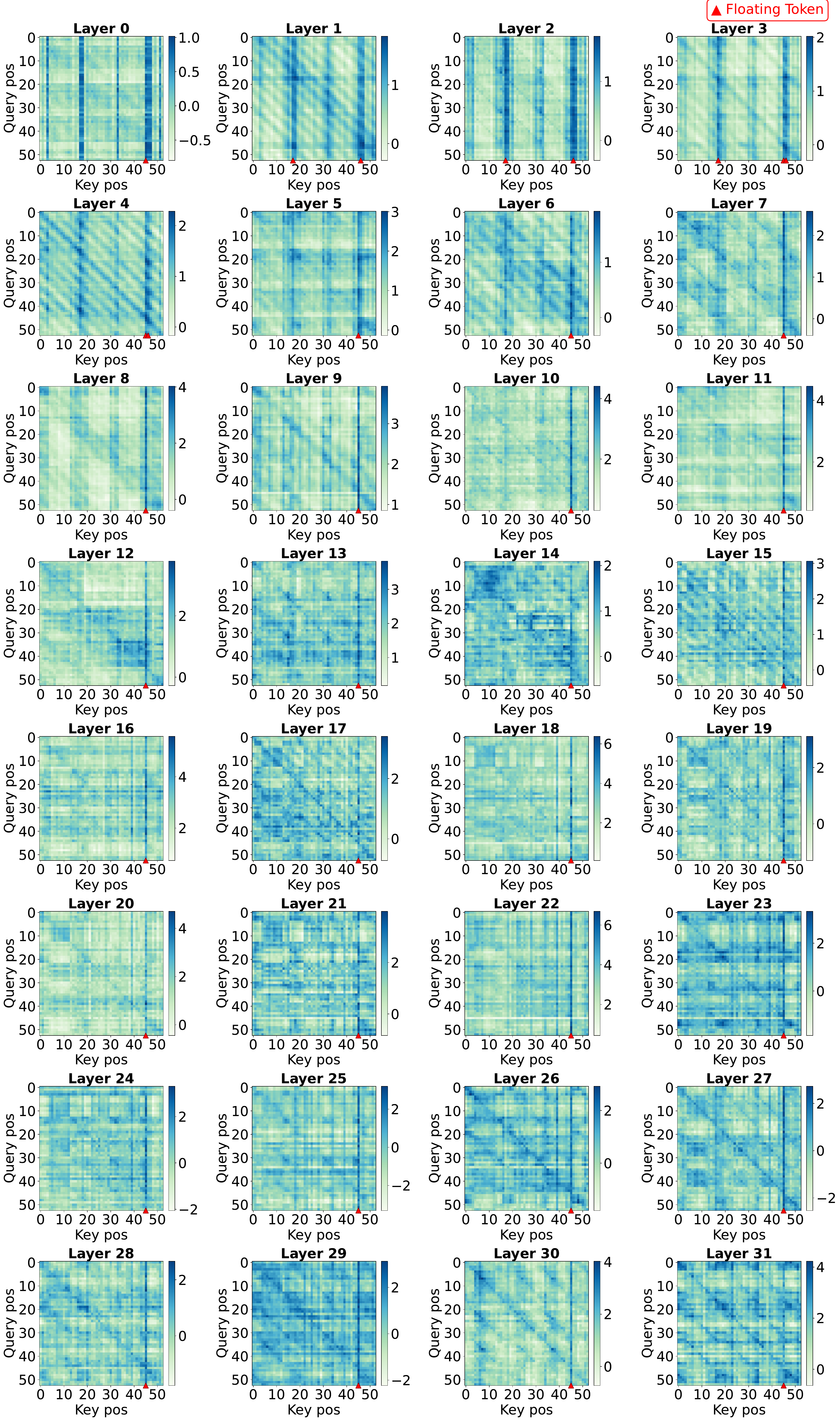}
    \caption{QK Score in Llada.}
    \label{fig:qk_fullseq_qk}
\end{figure*}

\begin{figure*}[t]
    \centering
    \includegraphics[height=0.90\textheight, keepaspectratio]{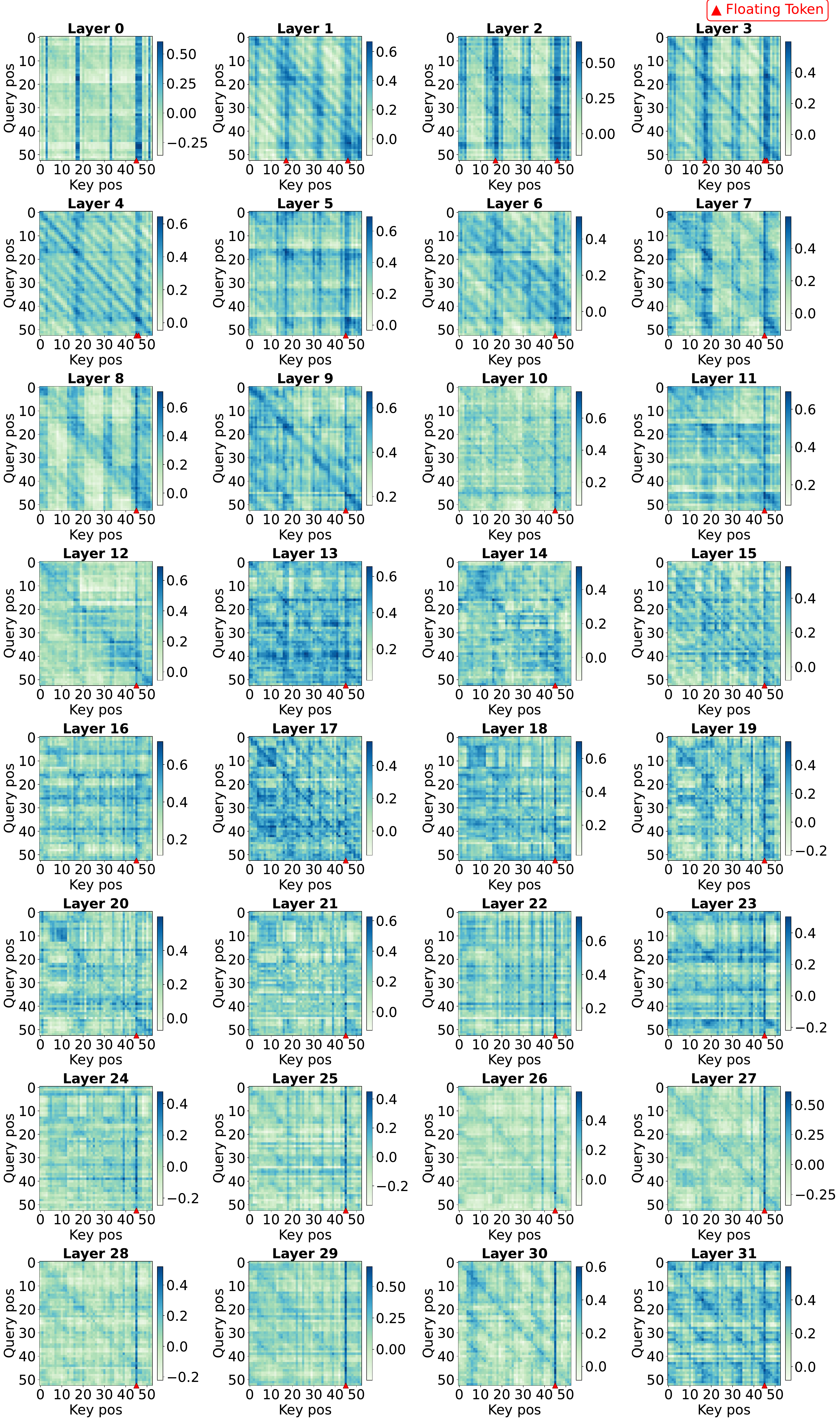}
    \caption{Angular in Llada.}
    \label{fig:qk_fullseq_cos}
\end{figure*}

\begin{figure*}[t]
    \centering
    \includegraphics[height=0.90\textheight, keepaspectratio]{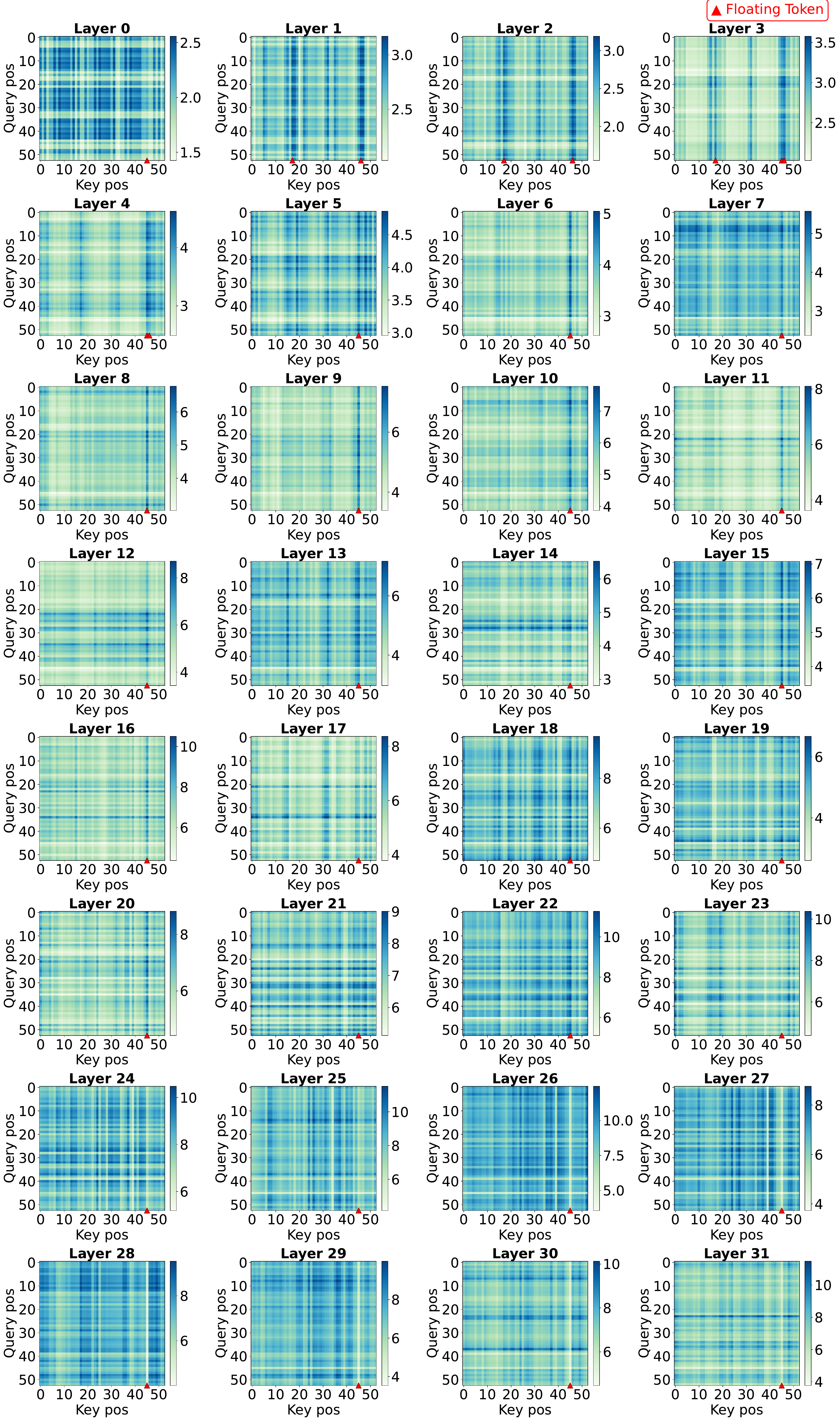}
    \caption{Norm Product in Llada.}
    \label{fig:qk_fullseq_normprod}
\end{figure*}

\begin{figure*}[t]
    \centering
    \includegraphics[height=0.90\textheight, keepaspectratio]{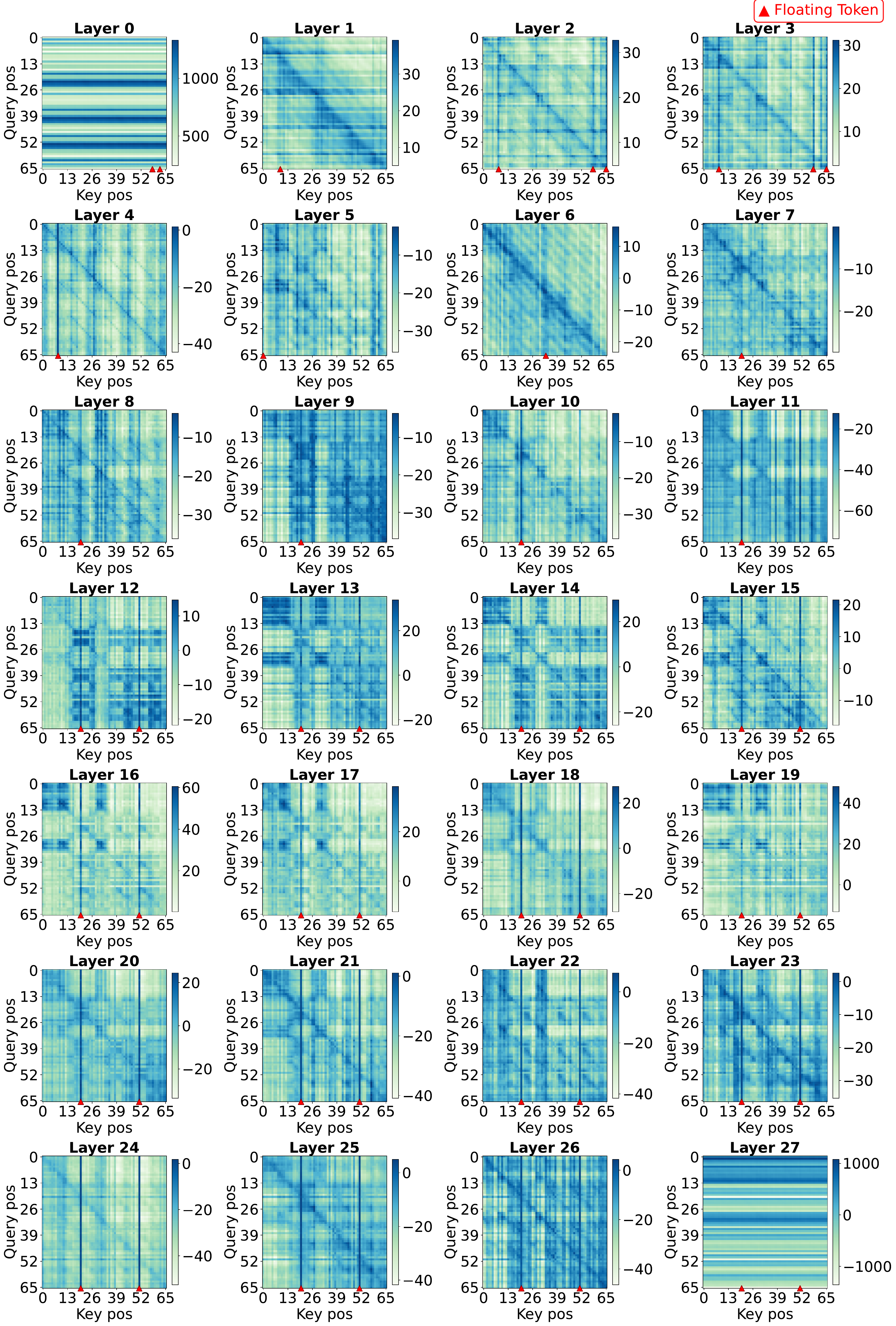}
    \caption{QK Score in Dream.}
    \label{fig:fullseq_qk}
\end{figure*}

\begin{figure*}[t]
    \centering
    \includegraphics[height=0.90\textheight, keepaspectratio]{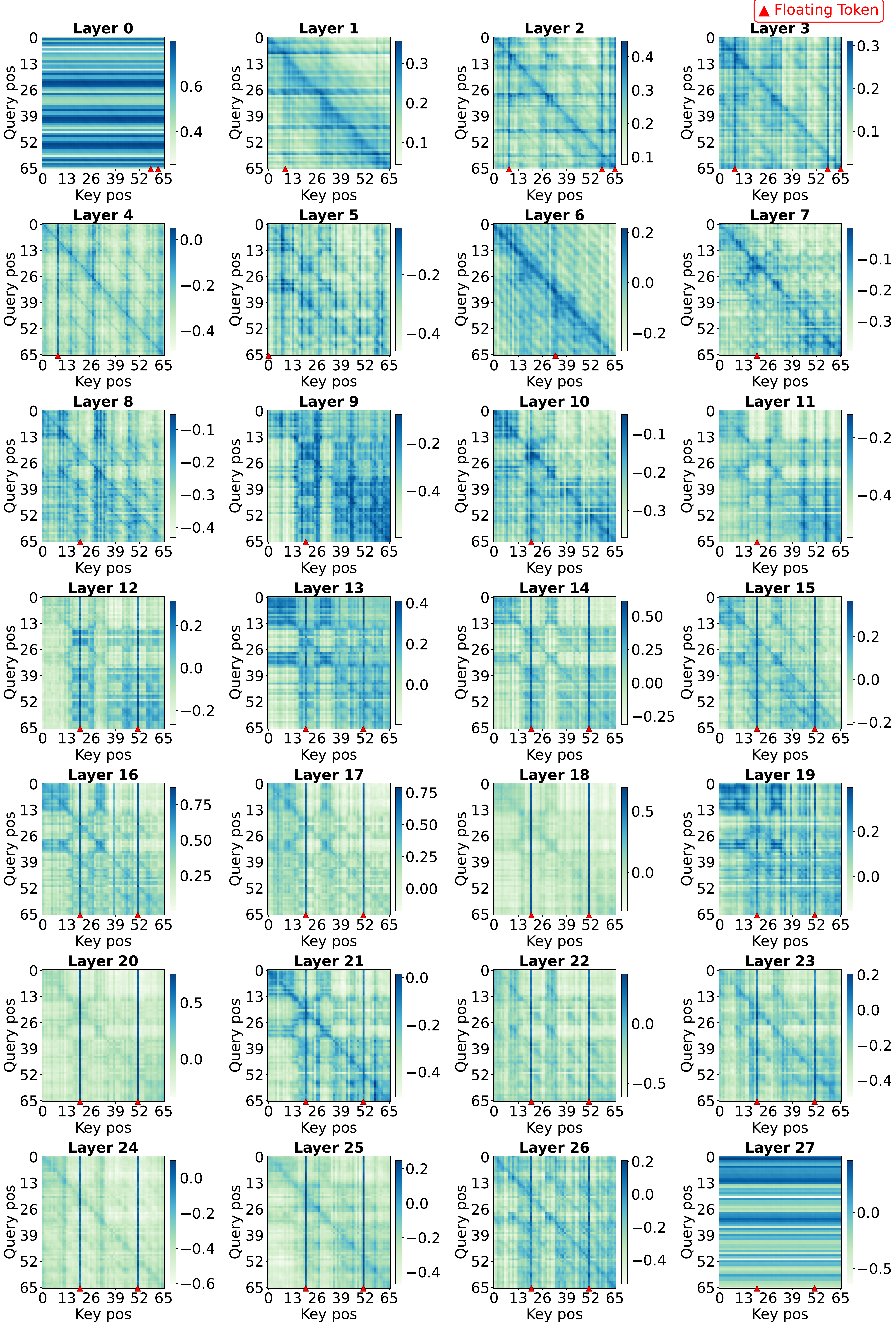}
    \caption{Angular in Dream.}
    \label{fig:fullseq_cos}
\end{figure*}

\begin{figure*}[t]
    \centering
    \includegraphics[height=0.90\textheight, keepaspectratio]{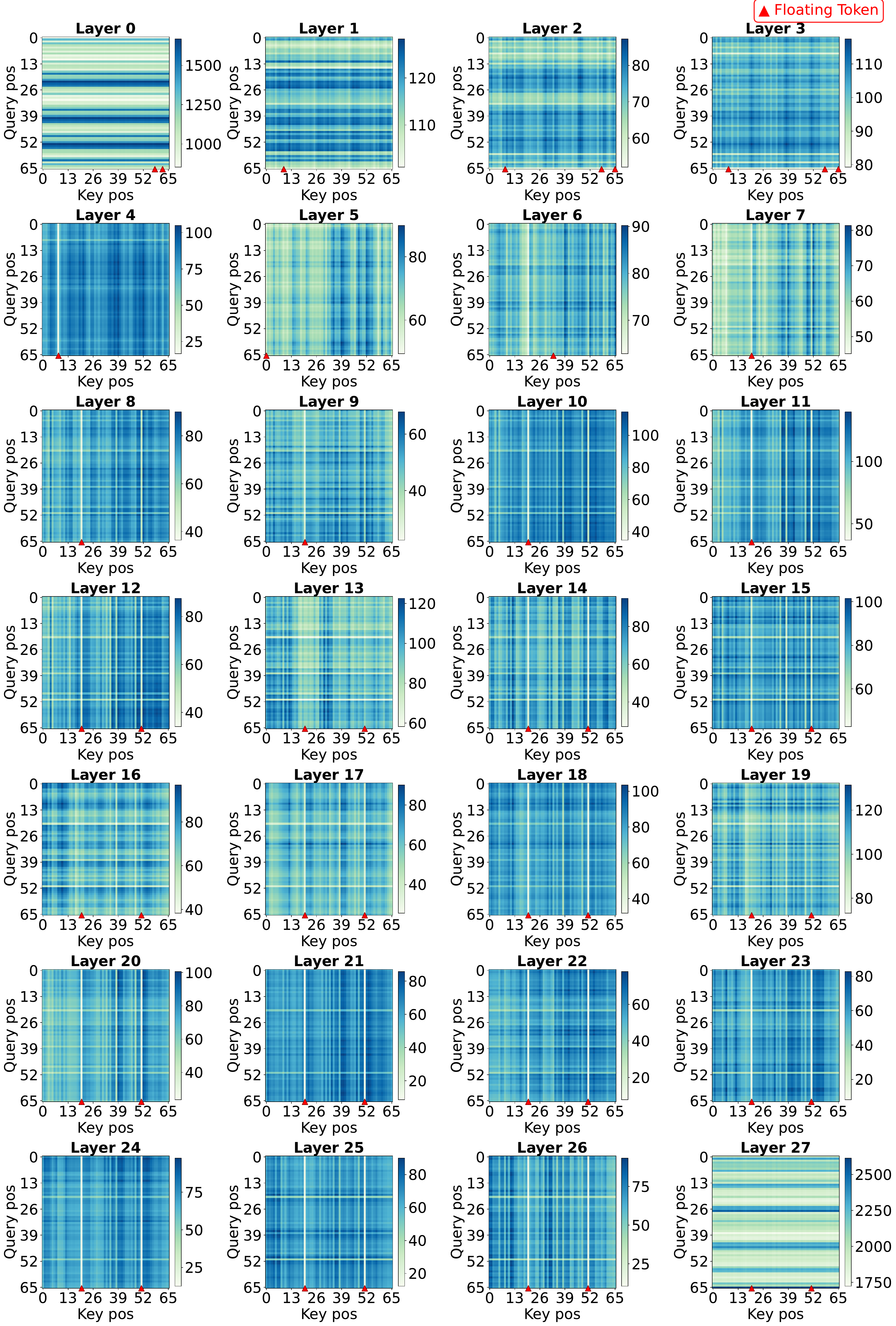}
    \caption{Norm Product in Dream.}
    \label{fig:fullseq_normprod}
\end{figure*}

\paragraph{Layer-wise QK Decomposition Heatmaps.}
\label{qk_appen}
We provide visualizations of the layer-wise QK decomposition across all layers, including the QK scores, angular and norm products in Figure~\ref{fig:qk_fullseq_qk}--Figure~\ref{fig:fullseq_normprod}, confirming that the proposed Shallow Structure-Aware, Deep Content-Focused attention mechanism is not an artifact of cropping or local windows.

\subsection{Attention Flow for Multi-Document RAG}
\label{app:rollout-multi-doc}

This section provides the formal definition of the attention flow procedure following~\cite{abnar2020quantifying} used in our analysis.

\begin{table}[t]
\centering
\small
\begin{tabular}{llcc}
\toprule
Task & Dataset & Total \\
\midrule
\multirow{3}{*}{Open-domain QA}
  & Natural Questions (2019) & 2{,}837 \\
  & TriviaQA (2017)          & 5{,}359 \\
  & MARCO QA (2016)          & 3{,}000 \\
\midrule
Multi-hop QA
  & HotpotQA (2018)          & 5{,}000 \\
\midrule
Slot Filling
  & T-REx (2018)             & 5{,}000 \\
\bottomrule
\end{tabular}
\caption{
Dataset statistics for Knowledge-Intensive Tasks.
}
\label{tab:data-stats}
\end{table}

\paragraph{Token-Level Influence Matrix.}
To account for residual connections in Transformers, we first augment the attention weights with a residual term:
\begin{equation}
\bar{A}^{\ell}_{i \rightarrow j} = \alpha A^{\ell}_{i \rightarrow j} + (1 - \alpha) \delta_{i \rightarrow j},
\end{equation}
where $\alpha$ is a hyperparameter controlling the balance between attention and residual connections, and $\delta_{i \rightarrow j}$ is the Kronecker delta. Following \citet{abnar2020quantifying}, we set $\alpha = 0.5$. 
Since adding the residual term changes the row sums, we re-normalize using the average attention received by each position:
\begin{equation}
\tilde{A}^{\ell}_{i \rightarrow j} = \frac{\bar{A}^{\ell}_{i \rightarrow j}}{\sum_{i=1}^n \bar{A}^{\ell}_{i \rightarrow j}},
\end{equation}
where $n$ denotes the length of the sequence.
We then accumulate attention across layers via matrix multiplication. Let $\tilde{\mathbf{A}}^{\ell} \in \mathbb{R}^{n \times n}$ denote the adjusted attention matrix at layer $\ell$. The token-level influence matrix is:
\begin{equation}
\mathbf{R} = \prod_{\ell=1}^{L} \tilde{\mathbf{A}}^{\ell}, 
\end{equation}
where $R_{i \rightarrow j}$ represents the cumulative flow of information from position $i$ to position $j$ across all layers.

\begin{table}[t]
\centering
\small
\setlength{\tabcolsep}{5pt}        % ↓ 缩小列间距
\renewcommand{\arraystretch}{1.05} % ↓ 略缩行高
\begin{tabular}{c p{3.0cm} c c}    % ↓ Token 列定宽，整体更窄
\toprule
\textbf{Rank} & \textbf{Token} & \textbf{Prop.} & \textbf{Type} \\
\midrule
1 & \texttt{\textbackslash n}        & 61.09\% & Structural \\
2 & \texttt{<|endoftext|>}           & 28.70\% & Structural \\
3 & \texttt{\textvisiblespace}       & 3.34\%  & Structural \\
4 & \texttt{<|mdm\_mask|>}           & 2.13\%  & Structural \\
5 & \texttt{,}                       & 1.23\%  & Structural \\
6 & \texttt{.}                       & 0.87\%  & Structural \\
7 & \texttt{)}                       & 0.53\%  & Structural \\
8 & \texttt{?}                       & 0.38\%  & Structural \\
9 & \texttt{the}                     & 0.24\%  & Lexical \\
\bottomrule
\end{tabular}
\caption{High-Frequency Floating Token Statistics.}
\label{tab:floating_tokens_freq}
\end{table}

\paragraph{Region-Level Influence Matrix.}
In the multi-document RAG setting, we partition the sequence into
contiguous regions corresponding to <BOS>, Query, Doc1–Doc10, and Answer.
Let $\mathcal{I}_p$ be the set of token indices belonging to region $p$
and $\mathcal{I}_q$ the set for region $q$.
We aggregate the token-level influences into a
region-level matrix $R^{\text{region}} \in \mathbb{R}^{P \times P}$ via:
\begin{equation}
    R^{\text{region}}_{p \rightarrow q}
    = \frac{1}{|\mathcal{I}_p|\,|\mathcal{I}_q|}
      \sum_{i \in \mathcal{I}_p}
      \sum_{j \in \mathcal{I}_q} R_{i \rightarrow j}.
\end{equation}
where $P$ is the number of regions.
For visualization, we row-normalize $R^{\text{region}}$ so that each row
represents the relative distribution of outgoing influence from a source
region to all target regions.

\begin{figure}[t]
    \centering
    % ---- 第一行 ----
    \begin{subfigure}[b]{0.48\linewidth}
        \centering
        \includegraphics[width=\linewidth]{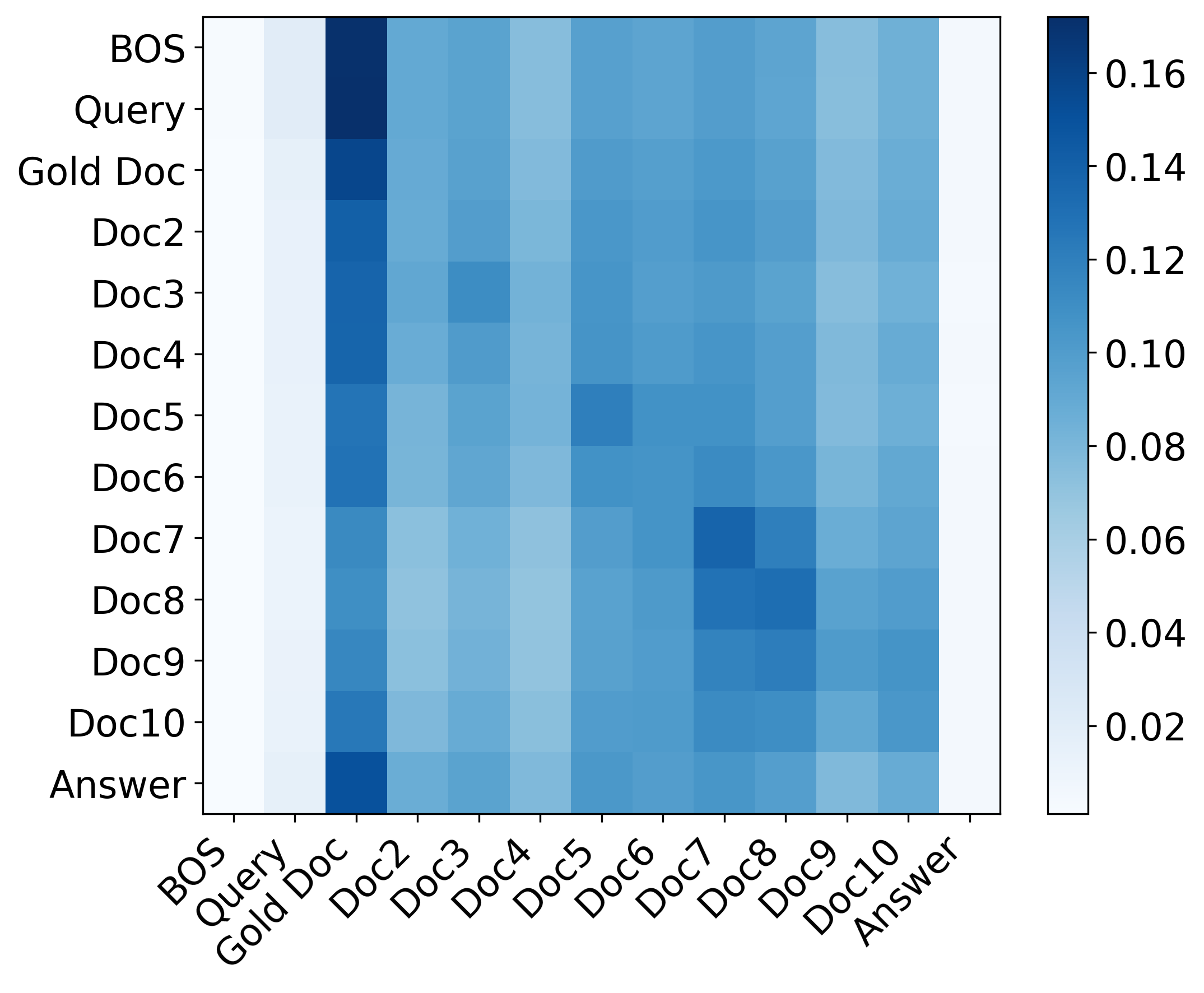}
        \caption{Dream with the Gold Doc at Position 1.}
        \label{fig:rollout-dream0}
    \end{subfigure}\hfill
    \begin{subfigure}[b]{0.48\linewidth}
        \centering
        \includegraphics[width=\linewidth]{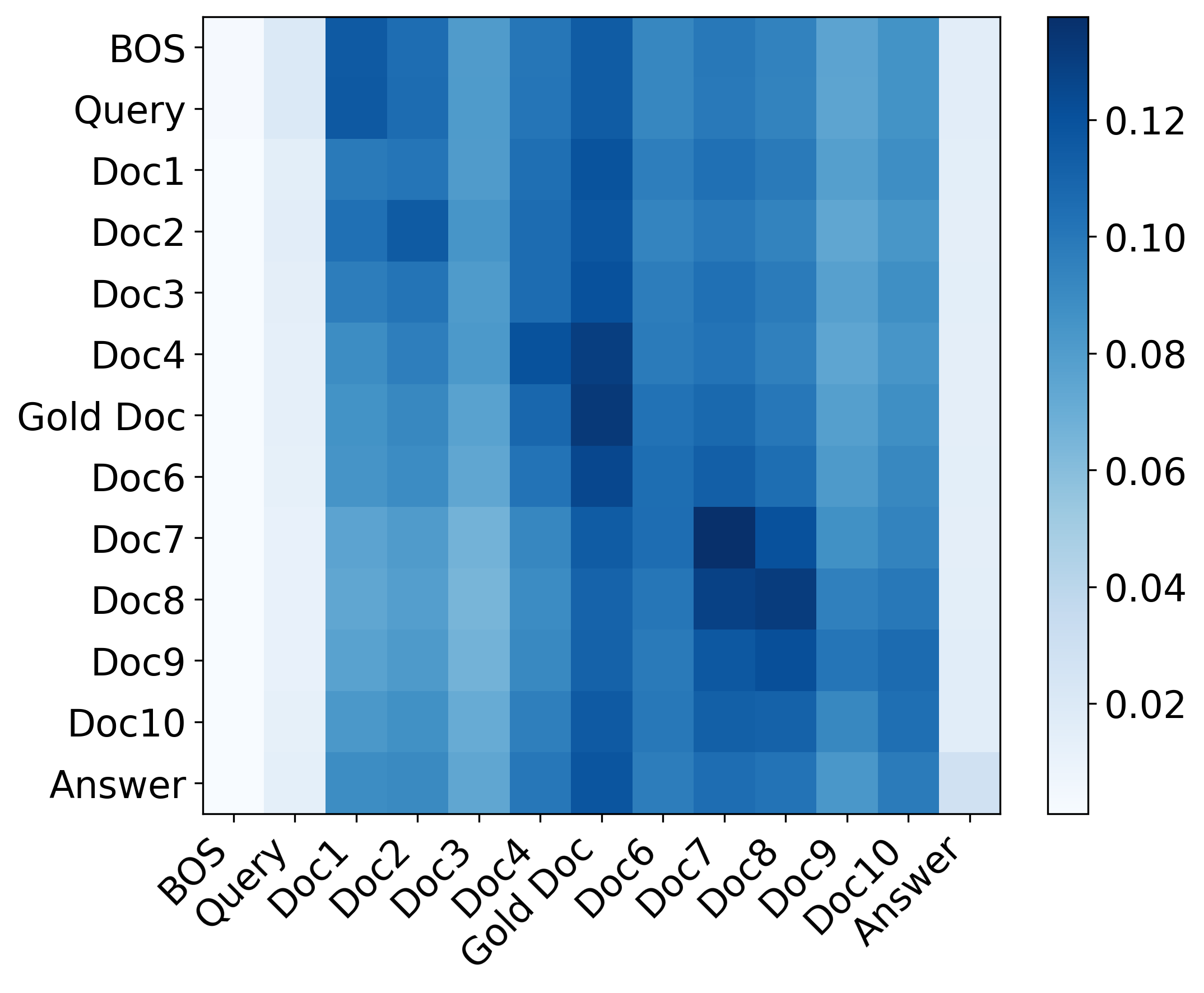}
        \caption{Dream with the Gold Doc at Position 5.}
        \label{fig:rollout-dream4}
    \end{subfigure}

    \vspace{0.5em}
    % ---- 第二行 ----
    \begin{subfigure}[b]{0.48\linewidth}
        \centering
        \includegraphics[width=\linewidth]{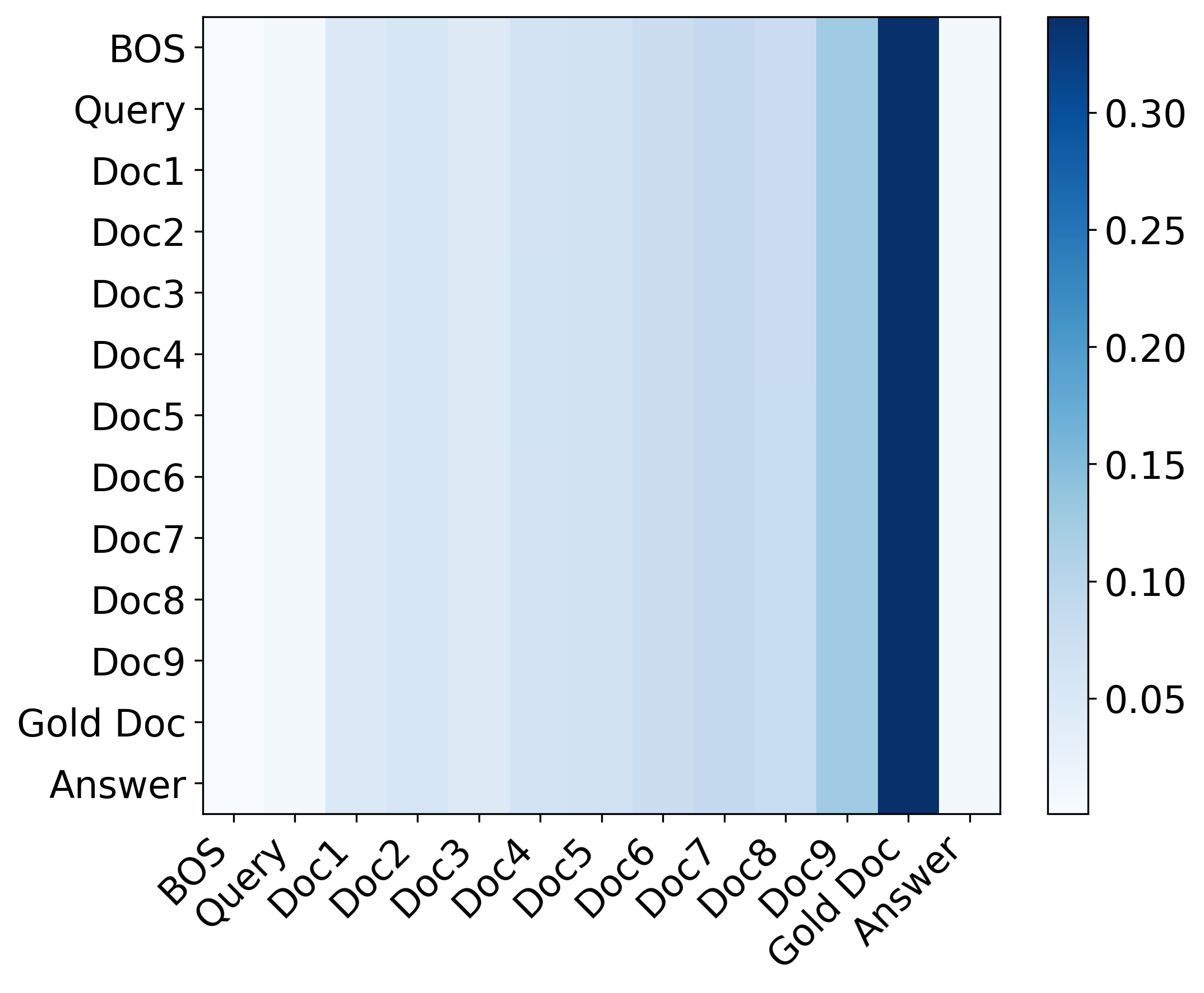}
        \caption{Dream with the Gold Doc at Position 10 .}
        \label{fig:rollout-dream9}
    \end{subfigure}\hfill
    \begin{subfigure}[b]{0.48\linewidth}
        \centering
        \includegraphics[width=\linewidth]{Figure/llama9.png}
        \caption{Qwen with Gold Doc at Positions 1, 5, 10.}
        \label{fig:rollout-qwen9}
    \end{subfigure}

    \caption{
    Region-Level Attention Flow for Dream (a–c) and Qwen (d) under Different Evidence Positions.}
    \label{fig:rollout-four-appendix}
\end{figure}

\paragraph{Region-Level Attention Flow for Other Models.}
\label{other region}
To verify that the observed positional behavior is consistent across model families, we additionally visualize the region-level attention flow for Dream (MDM) and Qwen (ARM) under different gold-document positions. As shown in Figure~\ref{fig:rollout-four-appendix}, Dream exhibits a clear shift of high-intensity flow toward the relocated evidence regions, while Qwen remains rigidly sunk in <BOS> region.

% (a) QK Score
\clearpage

% (b) Angular (Cos)
\clearpage

% (c) Norm Product
\clearpage

% (a) QK Score
\clearpage

% (b) Angular (Cos)
\clearpage 

% (c) Norm Product
\clearpage

\end{document}